\pdfoutput=1

\documentclass[10pt,journal,compsoc]{IEEEtran}

\usepackage[utf8x]{inputenc}
\usepackage{booktabs}
\usepackage{tablefootnote}
\usepackage{array}
\usepackage{graphicx}	% graphic
\usepackage{epstopdf}
\usepackage{float}
\usepackage{multirow}
\usepackage[colorlinks=true, citecolor=blue, linkcolor=green]{hyperref} % jump citation

\usepackage{soul}
\soulregister\cite7
\soulregister\ref7 
\usepackage{color, xcolor}
\soulregister{\cite}7 % 注册\cite命令
\soulregister{\ref}7 % 注册\ref命令
\soulregister{\pageref}7 % 注册\pageref命令
\soulregister{\caption}7

\usepackage{colortbl}

\ifCLASSOPTIONcompsoc
  \usepackage[nocompress]{cite}
\else
  \usepackage{cite}
\fi

\ifCLASSINFOpdf
\else
\fi

\begin{document}
	
\title{Towards Large-Scale Small Object Detection: Survey and Benchmarks}

%\author{Michael~Shell,~\IEEEmembership{Member,~IEEE,}
%        John~Doe,~\IEEEmembership{Fellow,~OSA,}
%        and~Jane~Doe,~\IEEEmembership{Life~Fellow,~IEEE}% <-this % stops a space
%\IEEEcompsocitemizethanks{
%\IEEEcompsocthanksitem M. Shell was with the Department
%of Electrical and Computer Engineering, Georgia Institute of Technology, Atlanta,
%GA, 30332.\protect\\
%note need leading \protect in front of \\ to get a newline within \thanks as
%\\ is fragile and will error, could use \hfil\break instead.
%E-mail: see http://www.michaelshell.org/contact.html
%\IEEEcompsocthanksitem J. Doe and J. Doe are with Anonymous University.}% <-this % stops an unwanted space
% \thanks{Manuscript received April 19, 2005; revised August 26, 2015.}}

\author{Gong~Cheng, Xiang~Yuan, Xiwen~Yao, Kebing~Yan, Qinghua~Zeng, Xingxing~Xie, and Junwei~Han,~\IEEEmembership{Fellow,~IEEE}
\IEEEcompsocitemizethanks{
	\IEEEcompsocthanksitem G. Cheng, X. Yuan, X. Yao, K. Yan, Q. Zeng, X. Xie and J. Han are with School of Automation, Northwestern Polytechnical University, Xi'an, 710021, China. Email: {gcheng, yaoxiwen}@nwpu.edu.cn, {shaunyuan, kebingyan, zengqinghua, xiexing}@mail.nwpu.edu.cn, junweihan2010@gmail.com
	\IEEEcompsocthanksitem Junwei Han is the corresponding author.}
}

\IEEEtitleabstractindextext{%
\begin{abstract}
With the rise of deep convolutional neural networks, object detection has achieved prominent advances in past years. However, such prosperity could not camouflage the unsatisfactory situation of Small Object Detection (SOD), one of the notoriously challenging tasks in computer vision, owing to the poor visual appearance and noisy representation caused by the intrinsic structure of small targets. In addition, large-scale dataset for benchmarking small object detection methods remains a bottleneck. In this paper, we first conduct a thorough review of small object detection. Then, to catalyze the development of SOD, we construct two large-scale Small Object Detection dAtasets (SODA), SODA-D and SODA-A, which focus on the Driving and Aerial scenarios respectively. SODA-D includes 24828 high-quality traffic images and 278433 instances of nine categories. For SODA-A, we harvest 2513 high resolution aerial images and annotate 872069 instances over nine classes. The proposed datasets, as we know, are the first-ever attempt to large-scale benchmarks with a vast collection of exhaustively annotated instances tailored for multi-category SOD. Finally, we evaluate the performance of mainstream methods on SODA. We expect the released benchmarks could facilitate the development of SOD and spawn more breakthroughs in this field. Datasets and codes are available at: \url{https://shaunyuan22.github.io/SODA}.
\end{abstract}

\begin{IEEEkeywords}
Object detection, Small object detection, Deep learning, Convolutional neural networks, Benchmark.
\end{IEEEkeywords}}

\maketitle

\IEEEdisplaynontitleabstractindextext

\IEEEpeerreviewmaketitle

\IEEEraisesectionheading{\section{Introduction}\label{sec:Sec1}}

\IEEEPARstart{O}{bject} detection is an essential task which aims at categorizing and locating the objects of interest in images/videos. Thanks to the enormous volume of data and powerful learning ability of deep Convolutional Neural Networks (CNNs), object detection has scored remarkable achievements in recent years \cite{1,2,3,4,5}. Small $ \footnote[1]{Here by “small” we mean the size of the object is relatively limited and often determined by an area \cite{6} or length \cite{7, 8} threshold.}$ Object Detection (SOD), as a sub-field of generic object detection, which concentrates on detecting those objects with small size, is of great theoretical and practical significance in various scenarios such as surveillance, drone scene analysis, pedestrian detection, traffic sign detection in autonomous driving, \textit{etc}. 

Albeit the substantial progresses have been made in generic object detection, the research of SOD proceeded at a relatively slow pace. To be more specific, there remains a huge performance gap in detecting small and normal sized objects even for leading detectors. Taking DyHead \cite{9}, one of the state-of-the-art detectors, as an example, the mean Average Precision (mAP) metric of small objects on COCO \cite{6} test-dev set obtained by DyHead is only $28.3\%$, significantly lag behind that of objects with medium and large sizes ($50.3\%$ and $57.5\%$ respectively). We posit such performance degradation originates the following two-fold: 1) the intrinsic difficulty of learning proper representation from limited and distorted information of small objects; 2) the scarcity of large-scale dataset for small object detection.

The low-quality feature representation of small objects can be attributed to their limited sizes and the generic feature extraction paradigm. Concretely, the current prevailing feature extractors \cite{10, 11, 12} usually down-sample the feature maps to diminish the spatial redundancy and learn high dimensional features, which unavoidably extinguishes the representation of tiny objects. Moreover, the features of small objects are inclined to be contaminated by background and other instances after the convolution process, making the network can hardly capture the discriminative information that is pivotal for the subsequent tasks. To tackle this problem, researchers have proposed a series of work, which can be categorized into six groups: sample-oriented methods, scale-aware methods, attention-based methods, feature-imitation methods, context-modeling methods, and focus-and-detect methods. We will discuss these approaches exhaustively in the review part and in-depth analyses will be provided too.

\begin{table*}[t]
	\tiny
	%\scriptsize
	\centering
	\caption{Summary of several surveys related to object detection. The top are the surveys focusing on the generic object detection and specific tasks, and the bottom are the existing reviews of small object detection.}
	\vspace{-1em}
	\resizebox{\textwidth}{!}{
		\begin{tabular}{p{0.42\textwidth} | m{0.055\textwidth} | p{0.51\textwidth}}	% {p{23.94em}|p{5.94em}|p{30.25em}}
			\toprule
			\textbf{Title} & \textbf{Publication}  & \textbf{Descriptions} \\
			\midrule
			Deep Learning for Generic Object Detection: A Survey \cite{13} & IJCV 2020 & A comprehensive survey of the recent progresses driven by deep learning techniques in generic object detection \\
			\midrule
			Object Detection With Deep Learning: A Review \cite{14} & TNNLS 2019 & A systematic review on deep learning-based detection frameworks for generic object detection and other subtasks \\
			\midrule
			Survey of Pedestrian Detection for Advanced Driver Assistance Systems \cite{15} & TPAMI 2009 & A survey focuses on pedestrian detection in advanced driver assistance systems \\
			\midrule
			Pedestrian detection: an evaluation of the state of the art \cite{16} & TPAMI 2011 & A detailed evaluation of pedestrian detectors in monocular images \\
			\midrule
			From Handcrafted to Deep Features for Pedestrian Detection: A Survey \cite{17} & TPAMI 2021 & A through survey for pedestrian detection approaches based on handcrafted features and deep features \\
			\midrule
			Text Detection and Recognition in Imagery: A Survey \cite{18} & TPAMI 2014 & A systematic survey related to automatic text detection and recognition in color images \\
			\midrule
			A survey on object detection in optical remote sensing images \cite{19} & JPRS 2016 & A review of recent progress about object detection in optical remote sensing images \\
			\midrule
			Object detection in optical remote sensing images: A survey and a new benchmark \cite{20} & JPRS 2020 & A thorough review of deep learning based methods for object detection in aerial images \\
			\midrule
			Vision for Looking at Traffic Lights: Issues, Survey, and Perspectives \cite{21} & TITS 2016 & An overview of traffic light recognition research in relation to driver assistance systems \\
			\midrule
			Object Detection Using Deep Learning Methods in Traffic Scenarios \cite{22} & CS 2021 & A survey dedicated to object detection in traffic scenarios based on deep learning methods  \\
			\midrule
			Imbalance Problems in Object Detection: A Review \cite{23} & TPAMI 2020 & A comprehensive review of the imbalance problems in object detection \\
			\midrule
			Weakly Supervised Object Localization and Detection: A Survey \cite{24} & TPAMI 2021 & A systematic survey on weakly supervised object localization and detection \\
			\bottomrule
		\end{tabular}%
		\label{tab:Tab1-1}%
	}
\end{table*}%
\begin{table*}[h!]
	\tiny
	\vspace{-2mm}
	\centering
	\resizebox{\textwidth}{!}{
		\begin{tabular}{p{0.42\textwidth} | m{0.055\textwidth} | p{0.51\textwidth}}	% {p{23.94em}|p{5.94em}|p{30.25em}}
			\toprule
			Deep learning-based detection from the perspective of small or tiny objects: A survey \cite{25} & IVC 2022 & A review of existing deep learning-based detection methods which can be utilized for small or tiny objects \\
			\midrule
			A survey and performance evaluation of deep learning methods for small object detection \cite{26} & ESWA 2021 & A survey of recently developed deep learning methods for small object detection \\
			\midrule
			A Survey of the Four Pillars for Small Object Detection: Multiscale Representation, Contextual Information, Super-Resolution, and Region Proposal \cite{27} & TSMCS 2022 & A review of small object detection based on four genres of techniques: multiscale representation, contextual information, super-resolution, and region-proposal \\
			\bottomrule
		\end{tabular}%
		\label{tab:Tab1-2}%
	}
\vspace{-2em}
\end{table*}%

To alleviate the data scarcity, some datasets tailored for small object detection have been proposed, \textit{e.g.}, SOD \cite{28} and TinyPerson \cite{7}. However, these small-scale datasets cannot meet the needs of training supervised CNN-based algorithms, which are “hungry” for a substantial amount of labeled data. In addition, several public datasets contain a considerable number of small objects, such as WiderFace \cite{8}, SeaPerson \cite{29} and DOTA$\footnote[2]{The term \textbf{DOTA} in our paper represents its 2.0 version, \textit{i.e.}, DOTA-v2.0.}$ \cite{30}, \textit{etc}. Unfortunately, these datasets are either designed for single-category detection task (face detection or pedestrian detection) which usually follows a relatively certain pattern, or among which tiny objects merely distribute in a few categories (\textit{small-vehicle} in DOTA dataset). In a nutshell, the currently available datasets could not support the training of deep learning-based models customized for small object detection, as well as serve as an impartial benchmark for evaluating multi-category SOD algorithms. Whilst, as a foundation for building data-driven deep CNN models, the accessibility of large-scale datasets such as PASCAL VOC \cite{31}, ImageNet \cite{32}, COCO \cite{6}, and DOTA \cite{30} is of great significance for both the academic and industrial communities, and each of which noticeably boosts the development of object detection in related fields. This inspires us to think: can we build a large-scale dataset, where the objects of multiple categories have very limited sizes, to serve as a benchmark that can be adopted to verify the design of small object detection framework and facilitate the further research of SOD? 

Taking the aforementioned problems into account, we construct two large-scale Small Object Detection dAtasets (SODA), SODA-D and SODA-A, which focus on the Driving and Aerial scenarios respectively. The proposed SODA-D is built on top of MVD \cite{33} and our data, where the former is a dataset dedicated to pixel-level understanding of street scenes, and the latter is mainly captured by on-board cameras and mobile phones. With 24828 well-chosen and high-quality images of driving scenarios, we annotate 278433 instances of nine categories with horizontal bounding boxes. SODA-A is the benchmark specialized for small object detection task under aerial scenes, which has 872069 instances with oriented rectangle box annotation across nine classes. It contains 2513 high-resolution images extracted from Google Earth.

\vspace{-1em}
\subsection{Comparisons with Previous Reviews}
Quite a number of surveys about object detection have been published in recent years \cite{13, 14, 15, 16, 17, 18, 19, 20, 21, 22, 23, 24}, and our review differs from the existing ones mainly in two aspects.

1. \textbf{A comprehensive and timely review dedicated to small object detection task across multiple domains}. Most of the previous reviews (as in Tab. \ref{tab:Tab1-1}) concentrate on either generic object detection \cite{13, 14} or specific object detection task such as pedestrian detection \cite{15, 16, 17}, text detection \cite{18}, detection in remote sensing images \cite{19, 20}, and detection under traffic scenarios \cite{21, 22}, \textit{etc}. Furthermore, there already exist several reviews paying their attention to small object detection \cite{25, 26, 27}, however, they either fail to the comprehensiveness and in-depth analysis because only partial reviews on limited areas were conducted, or categorize considerable algorithms belonging to generic detection as small object detection methods, which is indeed not rigorous for a SOD-oriented survey. By narrowly casting our sight to small/tiny objects, we extensively review hundreds of literature related to SOD task which covers a broad spectrum of research fields, including face detection, pedestrian detection, traffic sign detection, vehicle detection, object detection in aerial images, to name a few. As a result, \textbf{we provide a systematic survey of small object detection and an understandable and highly structured taxonomy, which organizes SOD approaches into six major categories based on the techniques involved and is radically different from previous ones}.

2. \textbf{Two large-scale benchmarks customized for small object detection were proposed, on which in-depth evaluation and analysis of several representative detection algorithms were performed}. Previous reviews mainly resort to general detection datasets such as PASCAL VOC \cite{31} and COCO \cite{6} to conduct evaluation, which is dominated by the medium-sized and large-sized instances and thereby failing to embody the authentic performance of related methods when it comes to small objects. Instead, we present the large-scale benchmark SODA and on top of which, a thorough evaluation of several representative generic object detection methods and newly published SOD approaches was provided.

\vspace{-1em}
\subsection{Scope}
Object detection in early period usually integrated handcrafted features \cite{34, 35, 36} and machine learning approaches \cite{37, 38} to recognize the objects of interest. The methods following this sophisticated philosophy perform catastrophically poorly in small objects due to their limited capability of scale variation. After 2012, the powerful learning ability of deep convolutional network \cite{39} brings a glimmer of hope to the whole detection community, especially considering that object detection had reached a plateau after 2010 \cite{40}. The seminal work \cite{40} broken the ice and since then, an increasing number of detection methods based on deep neural networks were proposed, whereafter, object detection entering the deep learning era. Thanks to the outstanding modeling ability of deep networks for scale variation and powerful abstraction of information, small object detection obtains an unprecedented improvement. Therefore, our review focuses on the major development of deep learning-based SOD methods. 

\begin{figure}[t]
	\centering
	\includegraphics[scale=0.6]{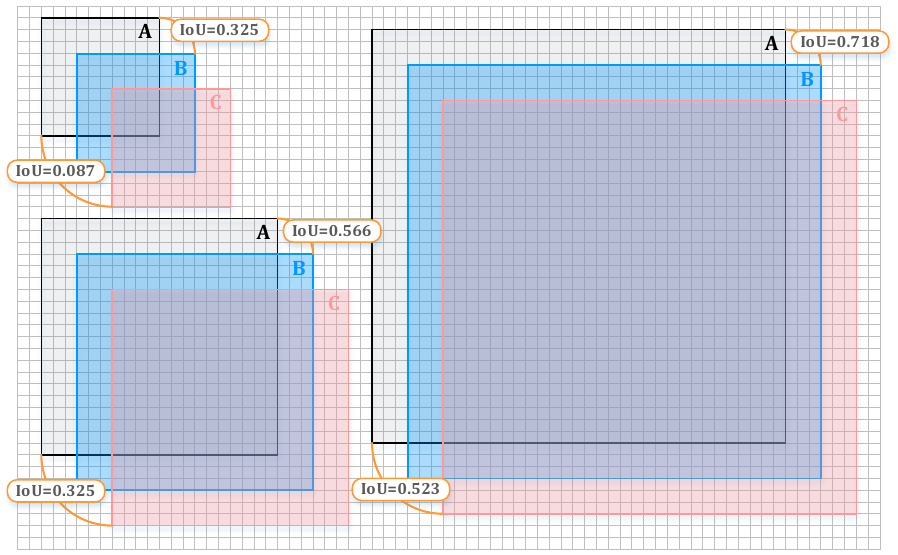} 
	\vspace{-1em}
	\caption{Low tolerance of small objects for bounding box perturbation. Top-left, bottom-left and right represent small object ($20\times 20$ pixels, a grid denotes two pixels), medium object ($40\times 40$ pixels) and large object ($70\times 70$ pixels), respectively. \textbf{A} denotes the Ground Truth (GT) box, \textbf{B} and \textbf{C} stand for predicted boxes with slight deviations along the diagonal direction ($6$ pixels and $12$ pixels, respectively). IoU indicates the Intersection-over-Union value between the GT box and the related predicted box.}
	\label{fig:Fig1}
\vspace{-1.5em}
\end{figure}

To sum up, the main contributions of this paper are in three folds:

1. Reviewing the development of small object detection in the deep-learning era and providing a systematic survey of the recent progress in this field, which can be grouped into six categories: sample-oriented methods, scale-aware methods, attention-based methods, feature-imitation methods, context-modeling methods, and focus-and-detect approaches. Except for the taxonomies, in-depth analysis about the pros and cons of these methods were also provided. Meanwhile, we review dozens of datasets that span over multiple areas which relate to small object detection.

2. Releasing two large-scale benchmarks for small object detection, where the first one was dedicated to driving scenarios and the other was specialized for aerial scenes. The proposed datasets are the first-ever attempt to large-scale benchmarks tailored for SOD. We hope these two exhaustively annotated benchmarks could help the researchers to develop and verify effective frameworks for SOD and facilitate more breakthroughs in this field. 

3. Investigating the performance of several representative object detection methods on our datasets, and providing in-depth analyses according to the quantitative and qualitative results, which could benefit the algorithm design of small object detection afterwards.

The remainder of this paper is organized as follows. In Section 2, we conduct a comprehensive survey of small object detection. And a thorough review on several publicly available benchmarks related to small object detection is given in Section 3. In Section 4, we elaborate the collection and annotation process, as well as the data characteristics, about the proposed benchmarks. In Section 5, the results and analyses of several representative methods on our benchmarks are provided. Finally, we conclude our work and discuss the prospective research directions of small object detection.

\begin{figure*}[h!]
	\centering
	%	\vspace*{-25mm}
	%	\includegraphics[width=0.7\linewidth]{algo_taxonomy}
	\includegraphics[scale=1.1]{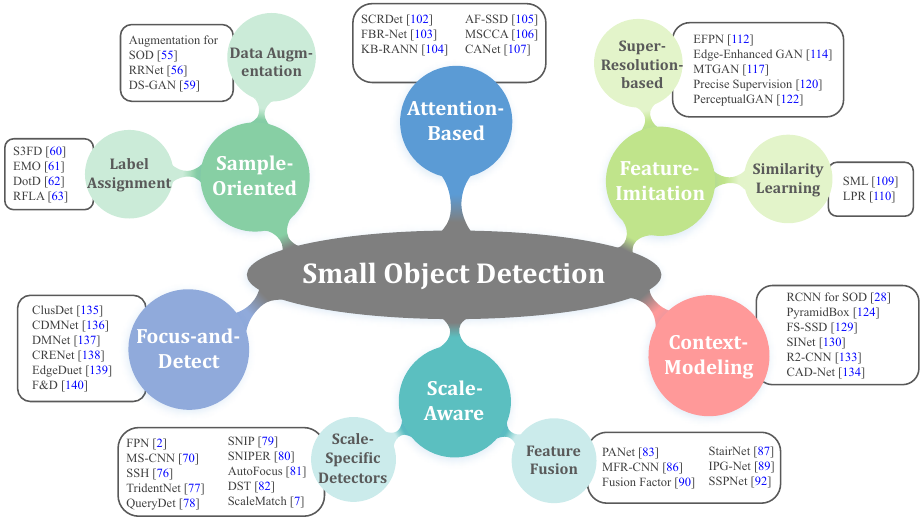}
	\vspace{-1em}
	\caption{Structured taxonomy of the existing deep learning-based methods for small object detection, which includes six genres. Only several representative methods of each category are demonstrated.}
	\label{fig:Fig2}
\vspace{-1em}
\end{figure*}

\section{Review on Small Object Detection}\label{sec:Sec2}
\subsection{Problem Definition}
Object detection aims to classify and locate instances. Small object detection or tiny object detection, as the term suggests, merely focus on detecting those objects with limited sizes. In this task, the terms \textit{tiny} and \textit{small} are typically defined by an area threshold \cite{6} or length threshold \cite{7, 8}. Take COCO \cite{6} as an example, the objects occupying an area less than and equal to 1024 pixels come to \textit{small} category. In this Section, we follow the expressions about those \textit{tiny} and \textit{small} terms in the original papers, and the definition of \textit{Small} in our benchmark will be introduced in Sec. \ref{sec:Sec4}.

\subsection{Main Challenges}\label{sec:Sec2.1}
In addition to some common challenges in generic object detection such as intra-class variations, inaccurate localization, occluded object detection, \textit{etc}., typical issues exist when it comes to SOD tasks, primarily including object information loss, noisy feature representation, low tolerance for bounding box perturbation and inadequate samples for training.

\textbf{Information loss}. Current prevailing object detectors \cite{1, 2, 3, 4, 5, 9} usually include a backbone network and a detection head, where the latter makes decision depends on the representation output by the former. Such paradigm was proven to be effective and gives rise to the unprecedented success. However, the generic feature extractor \cite{10, 11, 12} usually leverage sub-sampling operations to filter noisy activation \cite{41} and reduce the spatial resolution of feature maps, thus inevitably losing the information of objects. Such information loss will scarcely impair the performance of large or medium-sized objects to a certain extent, considering that the final features still retain enough information of them. Unfortunately, this is fatal for small objects, because the detection head can hardly give accurate predictions on top of the highly structural representations, in which the weak signals of small objects were almost wiped out.

\textbf{Noisy feature representation}. Discriminative features are crucial for both the classification and localization tasks \cite{42, 43}. Small objects often have low-resolution and poor-quality appearance, consequently it is intractable to learn representations with discrimination from their distorted structures. At the same time, the regional features of small objects are inclined to be contaminated by the background and other instances, introducing noise to the learned representation further. To sum up, the feature representations of small objects are apt to suffer from the noise, hindering the subsequent detection.

\textbf{Low tolerance for bounding box perturbation}. Localization, as one of the primary tasks of detection, is formulated as a regression problem in most detection paradigms, in which localization branch was designed to output the bounding box offsets \cite{1, 3, 44, 45, 46} or the object size \cite{4, 47}, and generally the Intersection over Union (IoU) metric was adopted to evaluate the accuracy. Nevertheless, localizing small objects is tougher than larger ones. As shown in Fig. \ref{fig:Fig1}, a slight deviation ($6$ pixels along the diagonal direction) of predicted box for a small object causes significant drop on IoU (from $100\%$ to $32.5\%$) compared to medium and large objects ($56.6\%$ and $71.8\%$). Meanwhile, a greater variance (say, $12$ pixels) further exacerbates the situation, and the IoU drops to poorly $8.7\%$ for small objects. That is to say, small objects have a lower tolerance for bounding box perturbation compared with larger ones, aggravating the learning of regression branch.

\textbf{Inadequate samples for training}. Selecting positive and negative samples is an indispensable step towards training a high performance detector. However, things get tougher when it comes to small objects. Concretely, small instances occupy fairly small regions and have limited overlaps to priors (anchors or points). This tremendously challenges conventional label assignment strategies \cite{1, 3, 4, 47, 50}, which collect \textit{pos/neg} samples based on the overlaps of boxes or center regions, leading insufficient positive samples assigned for small instances during training.

\vspace{-1em}
\subsection{Review of Small Object Detection Algorithms}
General object detection methods based on deep learning can be categorized into two groups: two-stage and one-stage detection, where the former detects objects in a coarse-to-fine routine while the later performs the detection at one stroke. Two-stage detection methods \cite{1, 46, 48} produce high-quality proposals with a well-designed architecture such as Region Proposal Network (RPN) \cite{1} at first, then the detection heads take regional features as input and perform subsequent classification and localization respectively. Compared with two-stage algorithms, one-stage approaches \cite{3, 44, 49} tile dense anchors on feature maps and predict the classification scores and coordinates directly. Benefiting from proposal-free setting, one-stage detectors enjoy high computational efficiency but often lag behind in accuracy. In addition to the above two categories, several anchor-free methods \cite{4, 47, 50, 51} have emerged in recent years, which discard the anchor paradigm. Moreover, query-based detectors \cite{5, 52}, which formulate the detection as a set prediction task, have shown great potential. We cannot elaborate on the related frameworks in the light of space restraints. Please refer to corresponding surveys \cite{13, 14} and original papers for more details.

To address the aforementioned challenging issues, existing small object detection methods usually introduce deliberate designs to current powerful paradigms which work well in generic object detection. Next, we will briefly introduce these approaches and an overview of the proposed solutions is presented in Fig. \ref{fig:Fig2}.

\subsubsection{Sample-oriented methods}\label{sec:Sec2.3.1}
One of the most critical procedures of training a learning-based detector is the sampling (often coexists with assignment), which has led to significant progress in generic object detection \cite{IQDet, PAA}. However, for SOD task, generic sampling strategies usually fail to provide adequate positive samples, thereby impairing the final performance. Such predicament originates from two aspects: the targets with limited sizes only occupy a small portion in current datasets \cite{6, 30, 31}; current overlap-based matching schemes \cite{1, 3, 4, 47, 50} are too rigorous to sample sufficient positive anchors or points owing to the limited overlaps between priors and the regions of small objects. In view of the two observations, a series of efforts have been made and can be split into two factions: increasing the number of small objects by data augmentation or devising optimal assignment strategy to enable adequate samples for network learning.

\textbf{Data-augmentation strategies}. Kisantal \textit{et al.} \cite{53} adopted an augmentation strategy by copying a small object and pasting it with random transformation to different positions in the identical image. RRNet \cite{54} introduces an adaptive augmentation strategy named AdaResampling, which follows the same philosophy as \cite{53}, the major difference lies in that a prior segmentation map was used to guide the sampling process of valid positions to be pasted, and a scale transformation for pasted objects reduces the scale discrepancy further. Zhang \textit{et al.} \cite{aug1} and Wang \textit{et al.} \cite{aug2} both employed divide-and-resize functionality-based operations to obtain more training samples of small objects. On top of the techniques of object segmentation, image inpainting and image blending, DS-GAN \cite{aug3} devises a novel data augmentation pipeline to generate high-quality synthetic data of small objects.

\textbf{Optimized label assignment}. Methods following this philosophy intend to alleviate the sub-optimal sampling result due to the overlap-based matching strategy and prior designs. With the help of the devised scale compensation anchor matching strategy, S$^3$FD \cite{S3FD} increases the matched anchors of tiny faces, thereby improving the recall rate. Zhu \textit{et al.} \cite{EMO} proposed Expected Max Overlapping (EMO) score, which takes anchor stride into account when computing the overlaps and enlightens better anchor setups for small faces. Xu \textit{et al.} \cite{136} employed the proposed DotD (defined as the Normalized Euclidean Distance between the center points of two bounding boxes) to replace the commonly used IoU. Similarly, RFLA \cite{137} measures the similarity between the Gaussian receptive field of each feature point and ground truth in label assignment, which boosts the performance of mainstream detectors on tiny objects.

Samples matter in object detection, especially for SOD task. Without enough positive samples, the regions of small objects are under-optimized during training and thereby hampering subsequent classification and regression. Either augmentation-based methods or devised matching strategies and appropriate prior settings intend to provide sufficient positive samples. Nevertheless, the former line of methods always suffers from inconsistent performance improvement and poor transferability. Meanwhile, current optimized label assignment schemes are prone to introduce low-quality samples and still struggle on the objects with extremely limited sizes.

\subsubsection{Scale-aware methods}
Objects in an image often vary in scale and such variation could be particularly severe in traffic scenarios and remote sensing images, leading disparate detection difficulties for a single detector. Previous approaches \cite{61, 62} usually employ image pyramid \cite{63} with sliding window scheme to handle the scale-variance issue. However, hand-crafted feature based methods, constrained by the limited representation capacity, perform catastrophically poorly on small objects. Early detection methods based on deep models still struggle in detecting tiny objects because only high-level features were used for recognition. To remedy the weakness of this paradigm and inspired by the success of reasoning across multi-level in other vision fields \cite{64, 65}, the following works mainly follow two paths. One refers to construct scale-specific detectors by devising multi-branch architecture or tailored training scheme, and the other line of efforts intends to fuse the hierarchical features for powerful representations of small objects.

\textbf{Scale-specific detectors}. The nature behind this line is simple: the features at different depths or levels were responsible for detecting the objects of corresponding scales only. Yang \textit{et al.} \cite{67} exploited scale-dependent pooling (SDP) to select a proper feature layer for subsequent pooling operation of small objects. MS-CNN \cite{68} generates object proposals at different intermediate layers, each of which focuses on the objects within certain scale ranges, enabling the optimal receptive field for small objects. Following this roadmap, DSFD \cite{69} employs two-shot detector connected by the feature enhancement module to detect the faces of various scales. YOLOv3 \cite{45} conducts multi-scale predictions by adding parallel branches where high-resolution features are responsible for small objects. Lin \textit{et al.} \cite{2} proposed Feature Pyramid Network (FPN), where the instances of various scales were assigned to different pyramid levels according to their sizes. Meanwhile, the interaction of features at different depths further guarantees the proper representation of multi-scale objects. This simple yet effective design has become an essential component in feature extractor and inspires a series of remarkable variants, \textit{e.g.}, NAS-FPN  \cite{NAS-FPN}, Bi-FPN \cite{Bi-FPN} , and Recursive-FPN \cite{R-FPN}. In addition, combining scale-wise detectors for multi-scale detection has been extensively explored. Li \textit{et al.} \cite{71} built parallel subnetworks where small-size subnetwork is learned specifically to detect small pedestrians. SSH \cite{72} combines scale-variant face detectors, each trained for a certain scale range, to form a strong multi-scale detector to handle the faces varying extremely in scales. TridentNet \cite{73} builds a parallel multi-branch architecture where each branch possesses optimal receptive fields for the objects of different scales. QueryDet \cite{76} designs the cascade query strategy to avoid the redundant computation on low-level features, making it possible to detect small objects on high-resolution feature maps efficiently.

Several approaches aim to develop tailored data preparation strategies to force the detector concentrate on the instances with specific scales during training. On top of generic multi-scale training scheme, Singh \textit{et al.} \cite{78} devised a novel training paradigm, Scale Normalization for Image Pyramids (SNIP), which only takes the instances whose resolutions fall into the desired scale range for training and the remainders are simply ignored. By this setting, small instances could be tackled at the most reasonable scales without compromising the detection performance on medium-to-large objects. Later, Sniper  \cite{79} advises to sample chips from a multi- scale image pyramid for efficient training. Najibi \textit{et al.} \cite{80} proposed a coarse-to-fine pipeline for detecting small objects. Considering that the collaboration between data preparation and model optimization is under-explored by previous methods  \cite{2, 63, 73}, Chen \textit{et al.} \cite{81} designed a feedback-driven training paradigm to dynamically direct data preparation and further balance the training loss of small objects. Yu \textit{et al.}  \cite{7} introduced a statistic-based match strategy for scale consistency.

\textbf{Hierarchical feature fusion}. Deep CNN architecture produces hierarchical feature maps at different spatial resolutions, in which low-level features describe finer details along with more localization cues, while high-level features capture richer semantic information \cite{13, 43, 73, 74, 82, 83}. For SOD task, deep features may struggle with the disappeared response of small objects, and the feature maps at early stages are susceptible to variations such as illumination, deformation and object pose, making the classification task more challenging. To overcome this dilemma, extensive approaches leverage feature fusion, which integrates the features at different depths, to obtain better feature representation for small objects. Enlightened by the simple-yet-effective interaction design in FPN \cite{2}, PANet \cite{74} enriches the feature hierarchy with bidirectional paths, enhancing deeper features with accurate localization signals. Zhang \textit{et al.} \cite{75} concatenated the pooled features of an RoI at multiple depths with global feature to obtain more robust and discriminative representation for small traffic objects. Woo \textit{et al.} \cite{86} proposed StairNet where deconvolution was exploited to enlarge the feature map, such learning-based up-sampling function can achieve a more refined feature than naive kernel-based up-sampling and allows that the information of different pyramid levels propagates more efficiently \cite{89}. Liu \textit{et al.} \cite{90} introduced IPG-Net, where a set of images at different resolutions obtained by the image pyramid  \cite{63} were input to the designed IPG transformation module to extract shallow features to complement spatial information and details. Gong \textit{et al.} \cite{87} devised a statistic-based fusion factor to control the information flow of adjacent layers. Noting that the gradient inconsistency encountered in FPN-based approaches deteriorates the representation ability of low-level features \cite{91}, SSPNet  \cite{88} highlights the features of specific scales at different layers and employs the relationship of adjacent layers in FPN to accomplish proper feature sharing. 

Scale-specific architectures are committed to processing small objects at most reasonable scale, and fusion-based approaches aim to bridge the spatial and semantic gaps between lower pyramidal levels and higher ones, both of them strive for the consistent performance gains of both small-scale objects and medium-to-large ones. However, the former maps the objects of different sizes to corresponding scale levels in a heuristic manner which may confuse the detectors, because the information of a single layer is inadequate to make accurate prediction. On the other hand, in-network information flow is not always conducive to the representations of small objects. Our goal is to not only endow the low-level features with more semantics, but also prevent the original responses of small objects from overwhelmed by deeper signals. Unfortunately, you cannot have your cake and eat it, hence this dilemma needs to be addressed carefully.

\subsubsection{Attention-based methods}
Human can quickly focus and distinguish objects while ignoring those unnecessary parts by a sequence of partial glimpses at the whole scene \cite{120}, and this astonishing capacity in our perception system is generally referred as visual attention mechanism, which plays a crucial role in our visual system \cite{123}. Not surprisingly, this powerful mechanism has been extensively investigated in the previous literature \cite{124, 125, 126, 127, 128} and shows great potential in many vision fields \cite{5, 9, 129, 130}. By allocating different weights to different parts of feature maps, the attention modeling indeed emphasizes the valuable regions while suppressing those dispensable ones. Naturally, one can deploy this superior scheme to highlight the small objects that are inclined to dominated by the background and noisy patterns in an image. 

SCRDet \cite{132} designs an oriented object detector, in which pixel attention and channel attention were trained in a supervised manner to highlight small object regions while eliminating the interference of noise. Extending the anchor-free detector FCOS \cite{4} with the proposed level-based attention, FBR-Net \cite{133} equilibrates the features at different pyramid levels and enhances the learning of small object under complicated situations. Enlightened by the human cognition, KB-RANN \cite{KB-RANN} exploits long-term and short-term attention neural networks to focus on the particular parts of image features, enhancing the detection of small objects. Lu \textit{et al.} \cite{AF-SSD} designed a dual path module to highlight the key feature of small objects and suppress the non-object information. By replacing the complex convolution components with the proposed enhanced channel attention (ECA) blocks, MSCCA \cite{MSCCA} constructs a lightweight detector with balanced channel features and less parameters. Li \textit{et al.} \cite{CANet} designed a cross-layer attention module to obtain stronger responses of small objects.

Drawing on the cognitive mechanism of mankind, visual attention plays an important role in nowadays vision fields, and it enables high-quality representations by screening the key parts while restraining noisy ones. Attention-series methods are highly claimed for their flexible embedding designs and can be plugged into almost all the SOD architectures, however, the performance improvement comes at the cost of heavy computation overhead owing to the correlation operations and moreover, current attention paradigms are lacking supervised signals and optimized implicitly.

\subsubsection{Feature-imitation methods}
One of the most significant challenges of SOD is the low-quality representations caused by the little information of small instances. This situation will likely get worse for those objects with extremely limited sizes \cite{94}. At the same time, larger instances often embody clear visual structures and better discrimination. Hence, a straightforward way to alleviate this low-quality issue is enriching the regional features of small objects by mimicking that of larger ones \cite{141}. To this end, several tentative methods have been proposed and can be categorized into two genres: feature imitation by similarity learning and super-resolution-based frameworks.

\textbf{Similarity learning-based methods}. The principle of this line is simple: imposing additional similarity constraints on the training of generic detectors, thereby bridging the representation gap between small objects and large ones. Wu \textit{et al.} \cite{141} proposed Self-Mimic Learning method, where the representations of small-scale pedestrians were enforced to approach to the local average RoI features of large-scale ones. Inspired by the memory process of human visual understanding mechanism, Kim \textit{et al.} \cite{142} devised a large-scale embedding learning with the large-scale pedestrian recalling memory (LPR Memory), and the overall architecture was optimized under the recalling loss which intends to guide the small- and large-scale pedestrian features to be similar.

\textbf{Super-resolution-based frameworks}. Methods following this roadmap aim at restoring the distorted structures of small objects instead of simply amplifying the ambiguous appearance of them. With the help of deconvolution and sub-pixel convolution \cite{97}, Zhou \textit{et al.} \cite{82} and Deng \textit{et al.} \cite{96} obtained high-resolution features specialized for small object detection. With self-supervised learning paradigm, Pan \textit{et al.} \cite{SSFA} proposed a guided feature upsampling module to learn upscaled feature representations with detailed information. Generative Adversarial Network (GAN) \cite{95} has remarkable capability to generate visually  authentic data by following a two-player minimax  game  between the generator and the discriminator, which, unsurprisingly, enlightens the researchers to explore this powerful paradigm for generating high-quality representations of small objects. Rabbi \textit{et al.} \cite{EE-GAN} and Bashir \textit{et al.} \cite{SRC-GAN} both use GAN to super-resolve low-resolution remote sensing images, where the former screens the edge details to avoid high-frequency information loss during reconstructing, and the latter incorporates the cyclic GAN and residual feature aggregation to capture complex features. Deeming that directly operating the whole images incurs non-negligible computational cost at feature extraction stage \cite{96}, MTGAN \cite{100} super- resolves the patches of RoIs with the generator network. Bai \textit{et al.} \cite{101} extended this paradigm to face detection task and Na \textit{et al.} \cite{102} applied super-resolution method to small candidate regions for better performance. Though super-resolving target patches could partly reconstruct the blurry appearance of small objects, this scheme neglects the contextual cues which play an important role for network prediction \cite{103, 104}. To deal with this issue, Li \textit{et al.} \cite{105} devised PerceptualGAN to mine and exploit the intrinsic correlations between small-scale and large-scale objects, in which the generator learns to map the weak representations of small objects to super-resolved ones to deceive the discriminator. To go a step further, Noh \textit{et al.} \cite{103} introduced direct supervision to the super-resolution procedure.

By adding additional similarity loss or super-resolution architectures to prevailing detectors, feature imitation methods empower the model to mine the intrinsic correlations between small-scale objects and large-scale ones, thereby enhancing the semantic representation of small objects. Nevertheless, either similarity learning-based methods or super-resolution-based approaches have to avoid the collapse problem and sustain the feature diversity. Moreover, GAN-based methods are inclined to fabricate spurious textures and artifacts, imposing negative impacts on detection. Worse still, the existence of super-resolution architecture complicates the end-to-end optimization.

\subsubsection{Context-modeling methods}
We human can effectively utilize the relationship between the environment and the objects or the relation of objects to facilitate the recognition of objects and scenes \cite{106, 107}. Such prior knowledge that captures the sematic or spatial associations is known as context, which conveys the evidence or cues beyond the object regions. The contextual information is of critical importance not only in visual systems of human \cite{104, 106}, but also in scene understanding tasks such as object recognition \cite{108}, semantic segmentation \cite{109} and instance segmentation \cite{110}, \textit{etc}. Interestingly, informative context sometimes can provide more decision support than the object itself, especially when it comes to recognizing the objects with poor viewing quality \cite{106}. To this end, several methods exploit the contextual cues to boost the detection of small objects.
 
Chen \textit{et al.} \cite{28} employed the representations of context regions which encompass the proposal patches for subsequent recognition. Hu \textit{et al.} \cite{92} investigated how to effectively encode the regions beyond the object extent and model the local context information in a scale-invariant manner to detect tiny faces. PyramidBox \cite{107} makes full use of contextual cues to find small and blur faces that are indistinguishable from background. The intrinsic correlations of objects in an image can be regarded as context likewise. FS-SSD \cite{111} exploits the implicit spatial context information, the distances between intra-class and inter-class instances, to redetect the objects with low confidences. Assuming that the original RoI pooling operation would break up the structures of small objects, SINet \cite{112} introduces a context-aware RoI pooling layer to maintain the contextual information. IONet \cite{113} computes global contextual features by two four-directional IRNN structures \cite{114} for better detection of small and heavily occluded objects. $\mathcal{R}^2$-CNN \cite{131} employs a global attention block to suppress false alarms and efficiently detect small objects in large-scale remote sensing images. Zhang \textit{et al.} \cite{CAD-Net} captured the correlations between objects and global scene (global context), as well as that between objects and their neighboring instances (local context) to improve the performance of small objects.

From the information theory perspective, the more types of features are considered, the more likely higher detection accuracy can be obtained \cite{75}. Inspired by the consensus, context priming has been extensively studied to generate more discriminative features, especially for small objects who have inadequate cues, enabling precise recognition. Unfortunately, both holistic context modeling or local context priming confuse about which regions should be encoded as context. In other words, current context modeling mechanisms determine the contextual regions in a heuristic and empirical fashion, which cannot guarantee the constructed representations are interpretable enough for detection.

\subsubsection{Focus-and-detect methods}
Small objects in high-resolution images tend to distribute non-uniformly and sparsely \cite{138}, and the general divide-and-detect scheme consumes too much computation on those empty patches, leading the inefficiency during inference. Can we filter out those regions with no object thereby reducing the useless operations to boost the detection? The answer is YES! Efforts in this area break the chain of generic pipeline for processing high-resolution images. They first abstract the regions contain targets, on which the detection performs subsequently.

Yang \textit{et al.} \cite{138} proposed a Clustered Detection network (ClusDet) that fully exploits the semantic and spatial information between objects to generate cluster chips and then performs the detection. Following this paradigm, Duan \textit{ et al.} \cite{57} and Li \textit{ et al.} \cite{139} both exploited pixel-wise supervision to density estimation, achieving more accurate density maps which characterize the distribution of objects well. CRENet \cite{CRENet} designs a clustering algorithm to adaptively search cluster regions. With tiling technique, Wang \textit{ et al.} \cite{EdgeDuet} developed EdgeDuet to enhance small object detection on edge devices. F$\&$S \cite{FS} introduces a Focus$\&$Detect framework, where Focusing Network detects candidate regions which then were cropped and resized to higher resolution, enabling the accurate detection of small objects. Deeming that the fixed-size input processing pipeline usually incurs missing detection of small objects, \cite{TILLING} exploits tilling method to detect pedestrians and vehicles in high-resolution aerial images in real time. 

Compared to generic sliding window mechanism, focus-and-detect methods empower adaptive crops and flexible zoom-in operation, \textit{i.e.}, smaller objects can be processed at higher resolutions while larger ones can be detected in a relatively lower resolution, which significantly saves memory footprint at inference and reduces the interference of background. Methods following this roadmap have to answer the key question: \textit{where to focus}? Current approaches resort to either manually additional annotations  or auxiliary architectures like segmentation network or Gaussian Mixture Model, yet the former requires laborious labeling while the latter complicates the end-to-end optimization.

\section{Review of Datasets for Small Object Detection}\label{sec:Sec3}
\subsection{Datasets for Small Object Detection}
Datasets are the cornerstone of learning-based object detection methods, especially for data-driven deep learning approaches. In the past decades, various research institutions have launched plenty of high-quality datasets \cite{6, 30, 31, 32}, and these publicly available benchmarks provide impartial platforms for validating the detection methods and significantly boost the development of related fields. Unfortunately, very few benchmarks are designed for small object detection. For the sake of integrity, we still retrospect a dozen datasets which contain considerable number of small objects, and expect to provide a comprehensive review of datasets. Instead of restricting our scope to specific tasks, we investigate the related datasets which span over a wide range of research areas, including face detection \cite{8}, pedestrian detection \cite{7, 115, 116}, object detection in aerial images \cite{20, 30, 118, 143}, to name a few. The statistics of these benchmarks are given in Tab. \ref{tab:Tab2}, and only the most representative among them were introduced below in detail due to the space restriction.

\textbf{COCO}. Pioneering works \cite{31, 32}, though push forward the development of vision recognition tasks, have been criticized for their ideal condition, where objects usually have large sizes and center on the images, bearing little resemblance to the real-world scenarios. To bridge this gap and foster fine level image understanding, COCO \cite{6} was launched in 2014, its trainval set annotates 886K objects distributed in 123K images with instance-level mask, covering 80 common categories under complex everyday scenes. Comparing to previous datasets for object detection, COCO contains more small objects (about $30\%$ instances in COCO trainset have an area less than 1024 pixels) and more densely packed instances, both of which challenge the detectors. Moreover, the fully segmented annotation and the reasonable evaluation metric encourage more accurate localization. All these features help COCO be the de facto standard for validating the effectiveness of object detection methods in past years.

\textbf{WiderFace}. WiderFace \cite{8} is a large-scale benchmark towards accurate face detection, in which faces vary significantly in scale, pose, occlusion, expression, appearance and illumination. It contains 32203 images with a total of 393703 instances. Except common bounding box annotations, attributes including occlusion, pose and event categories were also provided, which allows thorough investigation for existing approaches. The faces in WiderFace are divided into three subsets, namely small (between 10-50 pixels), medium (between 50-300 pixels) and large (larger than 300 pixels), where small subset accounts for half of all instances.

\textbf{TinyPerson}. TinyPerson \cite{7} focuses on the seaside pedestrian detection. TinyPerson annotates 72561 persons in 1610 images which are categorized into two subsets: tiny and small, according to their lengths. Due to the extremely tiny size, an ignore label was assigned to those regions that cannot be certainly recognized. As the first dataset dedicated to tiny-scale pedestrian detection, TinyPerson is a concrete step towards for tiny object detection. However, its limited number of instances and single pattern restrict its capacity to serve as a benchmark for SOD.

\textbf{TT100K}. TT100K \cite{117} is a dataset for realistic traffic sign detection which includes 30000 traffic sign instances in 100000 images, covering 45 common Chinese traffic-sign classes. Each sign in TT100K is annotated with precise bounding box and instance-level mask. The images in TT100K are captured from Tencent Street Views, holding a high degree of variability in weather conditions and illumination. Moreover, TT100K contains considerable small instances ($80\%$ of instances occupy less than $0.1\%$ in the whole image area) and the entire dataset follows a long-tail distribution. 

\textbf{VisDrone}. VisDrone \cite{147} is a large-scale drone-captured dataset which is collected over various urban/suburban areas of 14 different cities across China. Concentrating on two essential tasks in computer vision, VisDrone supports four tracks: image object detection, video object detection, single object tracking and multi-object tracking. For image object detection track, there are 10209 images with a resolution of $2000 \times 1500$ pixels and 542K instances covering 10 common object categories in traffic scenarios. The images in VisDrone are captured with drones from various urban scenes, thereby containing a mass of small objects due to viewpoint variations and heavy occlusions.

\begin{table*}[t!]
	\scriptsize
	\centering
	\caption{Statistics of some benchmarks available for small object detection. ODNI stands for object detection in natural images and ODAI denotes object detection in aerial images. (1K = 1000, 1M = 1000K).}
	\vspace{-1em}
	\resizebox{\textwidth}{!}{
		% for scriptsize font
		\begin{tabular}{p{0.14\textwidth} | p{0.13\textwidth} | p{0.08\textwidth} | p{0.05\textwidth} | p{0.07\textwidth} | p{0.50\textwidth}}	
			\toprule
			\textbf{Dataset name} & \textbf{Task field} & \textbf{Publication}  & \textbf{\#Images}  & \textbf{\#Instances}  & \textbf{Descriptions and Characteristics} \\
			\midrule
			COCO \cite{6} & ODNI  & ECCV 2014 & 123K  & 886K  & One of the most popular datasets for generic object detection \\
			\midrule
			SOD \cite{28} & ODNI  & ACCV 2016 & 4925  & 8393  & A small-scale dataset for small object detection \\
			\midrule
			WiderFace \cite{8} & Face detection & CVPR 2016 & 32K   & 393K  & A large-scale benchmark with rich annotations for face detection \\
			\midrule
			EuroCity Persons \cite{115} & Pedestrian detection & TPAMI 2019 & 47K  & 219K  & The largest dataset for pedestrian detection captured from dozens of Europe cities \\
			\midrule
			WiderPerson \cite{116} & Pedestrian detection & TMM 2020 & 13K   & 39K   & Pedestrian detection benchmark in traffic scenarios \\
			\midrule
			TinyPerson \cite{7} & Pedestrian detection & WACV 2020 & 1610  & 72K   & The first dataset dedicated to tiny-scale pedestrian detection \\
			\midrule
			STS Dataset \cite{STS} & Traffic sign detection & SCIA 2011 & 20000 & 3488 & The first publicly available traffic sign dataset for detection \\
			\midrule
			LISA \cite{LISA} & Traffic sign detection & TITS 2012 & 6610 & 7855 & A traffic sign dataset allowing for detection and tracking \\
			\midrule
			GTSDB \cite{GTSDB} & Traffic sign detection & IJCNN 2013 & 900 & 1206 & A benchmark for traffic sign detection collected under different scenarios \\
			\midrule
			TT100K \cite{117} & Traffic sign detection & CVPR 2016 & 100K  & 30K   & A realistic and large-scale benchmark for traffic sign detection \\
			\midrule
			BSTLD \cite{BSTLD} & Traffic light detection & ICRA 2017 & 13427 & 24000 & A large dataset for detecting traffic lights whose sizes down to 1 pixel in width \\
			\midrule
			UCAS-AOD \cite{UCAS-AOD} & ODAI & ICIP 2015 & 910 & 6029 & A aerial dataset collected from Google Earth for detection \\
			\midrule
			VEDAI \cite{VEDAI} & ODAI & JVC 2016 & 1268 & 2950 & A database dedicated to small vehicle detection in aerial images \\
			\midrule
			xView \cite{155} & ODAI & arXiv 2018 & 	1128 & 1M & One of the largest and most diverse available dataset of overhead imagery \\
			\midrule
			DIOR \cite{20} & ODAI  & JPRS 2020 & 23K   & 192K  & One of the most frequently used benchmarks for object detection in aerial images \\
			\midrule
			UAVDT \cite{150} & ODAI & IJCV 2020 & 80K & 841K & A dataset collected by Unmanned Aerial Vehicles for object detection and tracking \\
			\midrule
			VisDrone \cite{147} & ODAI & TPAMI 2021 & 189K & 2.5M & A large-scale drone-captured benchmark for detection and tracking \\
			\midrule
			DOTA \cite{30} & ODAI  & TPAMI 2021 & 11K   & 1.79M & The largest remote sensing detection dataset including considerable small objects \\
			\midrule
			AI-TOD$\tablefootnote[3]{The term \textbf{AI-TOD} in our paper denotes the latest version, \textit{i.e.}, AI-TOD-v2.}$ \cite{118} & ODAI  & JPRS 2022 & 28K   & 700K  & A tiny object detection dataset based on previous available datasets \\
			\midrule
			NWPU-Crowd \cite{119} & Crowd counting & TPAMI 2021 & 5109  & 2.13M & The largest dataset for crowd counting and localization to date \\
			\bottomrule
		\end{tabular}%
		\label{tab:Tab2}%
	}
	\vspace{-1em}
\end{table*}%

\begin{table}[t!]
	\vspace{-1em}
	\centering
	\caption{Area subsets and corresponding area ranges of objects in SODA benchmark.}
	\vspace{-1em}
	\resizebox{\columnwidth}{!}{
		\begin{tabular}{ccccc}
			\toprule
			\multirow{2}[4]{*}{\textbf{Area Subset}} & \multicolumn{3}{c}{\textit{Small}} & \multirow{2}[4]{*}{\textit{Normal}} \\
			\cmidrule{2-4} & \textit{extremely Small} & \textit{relatively Small} & \textit{generally Small} &  \\
			\midrule
			\textbf{Area Range} & $\left( 0,144 \right]$ &  $\left( 144,400 \right]$ &  $\left( 400,1024 \right]$ &  $\left( 1024,2000 \right]$ \\
			\bottomrule
		\end{tabular}%
		\label{tab:Tab3}%
	}
	\vspace{-1em}
\end{table}%

\textbf{DOTA}. DOTA \cite{30} is proposed to facilitate the object detection in Earth Vision. It contains 18 common categories and 1793658 instances in 11268 images. Each object has been annotated with horizontal/oriented bounding box. Owing to the high diversity of orientations in overhead view images and large-scale variations among instances, DOTA dataset has numerous small objects, but they only distribute in a few categories (\textit{small-vehicle}).

\subsection{Evaluation Metrics}
Before diving into the evaluation criteria of small object detection, we first introduce related preliminary concepts.
Given a ground-truth bounding box $b_{\rm g}$ and a predicted  box $b_{\rm p}$ output by the detector, if the IoU between $b_{\rm g}$ and $b_{\rm p}$ is greater than the predefined threshold, and the predicted label is in accordance with the ground-truth, the current detected box will be identified as a potential prediction to this object, also known as True Positive (TP),  otherwise it will be regarded as a False Positive (FP). Once we obtain the number of TP, FP and False Negative (FN, also known as missed positives), the Average Precision (AP) can be computed to evaluate the performance of detectors.

\textbf{Average Precision}. Average Precision (AP) is originally introduced in VOC2007 Challenge \cite{31} and usually adopted in a category-wise manner. Concretely, given a confidence threshold and an IoU threshold $\beta$ (0.5 for VOC2007), the Recall (\textit{R}) and Precision (\textit{P}) can be calculated afterwards. By varying the confidence threshold $\alpha$, one can obtain different pairs $(P, R)$ and ultimately, AP can be determined by averaging the precision scores under different recalls. This fixed IoU based AP metric once dominated the community for years.

\begin{figure*}[t!]
	\centering
	\includegraphics[scale=0.16]{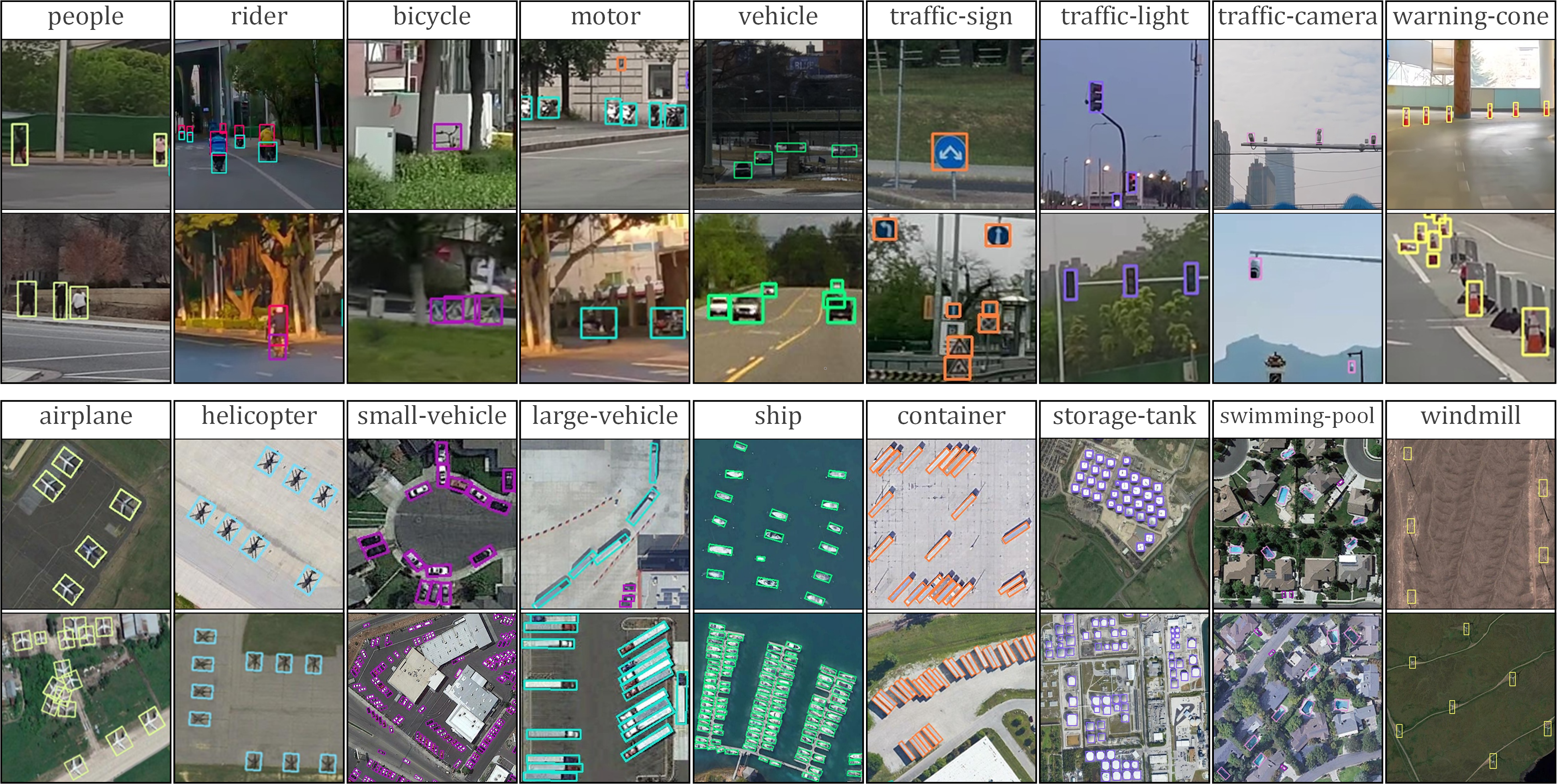}
	\vspace{-1em}
	\caption{Example instances of each category in SODA-D (Top) and SODA-A (Bottom).}
	\label{fig:Fig4}
	\vspace{-1em}
\end{figure*}

A new evaluation metric was introduced with the launch of COCO dataset after 2014, which averages AP across multiple IoU thresholds between 0.5 and 0.95 (with an interval of 0.05). Apart from merely considering fixed IoU threshold, this criterion also takes the higher IoU thresholds into account, encouraging more accurate localization. This reasonable evaluation metric has been used as the “gold standard” in detection community and widely adopted by the following works \cite{117, 144}. Noting that the overall AP is computed by averaging the APs of all categories in practice.

\section{Benchmarks}\label{sec:Sec4}
In this section, we briefly introduce the data acquisition and annotation process for building SODA-D and SODA-A. Then, we shed light on the characteristics of our benchmarks and the main differences between our datasets and related existing ones. Moreover, other details such as scene selection, data cleaning and annotation principles will be discussed in the Sec. \ref{app:bench} of Appendix.

\subsection{Data Acquisition and Annotation}\label{sec:Sec4.1}
Our aim is to build datasets tailored for small object detection, hence the point is \textbf{how to define a valuable object}.

\begin{table}[t!]
	\centering
	\caption{Numbers of instances of each category and three splits of SODA-D (Left) and SODA-A (Right).}
	\resizebox{0.95\columnwidth}{!}{
		\begin{tabular}{cc|cc}
			\toprule
			\textbf{Category} & \textbf{\#Instances} & \textbf{Category} & \textbf{\#Instances} \\
			\midrule
			people & 35928 & airplane & 31622 \\
			rider & 4636  & helicopter & 1395 \\
			bicycle & 2560  & small-vehicle & 526047 \\
			motor & 3896  & large-vehicle & 17006 \\
			vehicle & 69197 & ship  & 65690 \\
			traffic-sign & 85905 & container & 138242 \\
			traffic-light & 62729 & storage-tank & 35331 \\
			traffic-camera & 7636  & swimming-pool & 29735 \\
			warning-cone & 5946  & windmill & 27001 \\
			\midrule
			Train & 134301 & Train & 344228 \\
			Validation & 56050 & Validation & 231439 \\
			Test  & 88082 & Test  & 296402 \\
			Total & 278433 & Total & 872069 \\
			\bottomrule
		\end{tabular}%
		\label{tab:Tab4}%
	}
	\vspace{-1.5em}
\end{table}%

\textbf{Definition about a valuable object}. Generally, a bounding box $B$ can be represented as $(x, y, w, h, \theta)$, where $(x, y)$ denotes the center location and $(w, h)$ indicates the width and height of the box respectively, the parameter $\theta$ stands for the orientation angle and is unused for horizontal annotation. Moreover, we use $S=w\times h$ to denote the pixel area of an object.
In line with the definition of small or tiny objects in previous works \cite{6, 7, 118}, we adopt the absolute area criterion and regard an instance who has an area smaller than $1024$ pixels, \textit{i.e.}, $S\leq 1024$, as a \textit{Small} object. Meanwhile, an object whose area between $1024$ and $2000$ pixels will be annotated as a \textit{Normal} object. Otherwise, the object comes to the \textit{ignore} category and will not influence the final evaluation results. Considering the detection difficulty increases sharply when the object size gets smaller, we further divide the \textit{Small} objects into three subsets: \textit{extremely Small} (\textit{eS}), \textit{relatively Small} (\textit{rS}) and \textit{generally Small} (\textit{gS}), as demonstrated in Tab. \ref{tab:Tab3}.

\textbf{Data source}. The images in SODA-D are mainly from MVD \cite{33}, self-shooting and the Internet. MVD is a large-scale dataset for semantic understanding of street scenes, of which 25000 high-quality images are captured from road views, highways, rural areas and off-road. Thanks to the high-quality and high-resolution property with MVD, we can obtain a large set of valuable instances with clear visual structure. For self-shooting part, we use on-board cameras and mobile phones to collect images of typical driving scenes in several Chinese cities, including Beijing, Shenzhen, Shanghai, Xi’an, Qingdao, Guangzhou, \textit{etc}. In addition, we also crawl images by searching keywords on the image search engines (Google, Bing, Baidu, \textit{etc}.). Finally, we obtained 24828 images of traffic scene.

Enlightened by the pioneering works \cite{20, 30}, Google Earth$ \footnote[4]{\url{https://earth.google.com/}}$ was leveraged to collect images for SODA-A, we extract 2513 images from hundreds of cities around the world suggested by the experts. It is noting that numerous images with cluttered background and high density which are closer to realistic challenges are captured. In addition, the images in SODA-A have a relatively high resolution and most of them enjoy a resolution larger than $4700\times 2700$, enabling the finer details and adequate context that are of great significance to small object detection \cite{106, 107}.

\begin{table*}[!t]
	\scriptsize
	\centering
	\caption{Comparisons between SODA-D and several related detection datasets under driving scene (Top), likewise for SODA-A and some detection datasets under aerial scenario (Bottom). Note that \textit{eS}, \textit{rS} and \textit{gS} stand for \textit{extremely Small}, \textit{relatively Small} and \textit{generally Small} according to our definition (see Tab. \ref{tab:Tab3}). And for each dataset, we only count the subsets whose annotations are available, see Split column. Avg. Res. denotes the average image resolution of the dataset. HBB/OBB denotes horizontal/oriented bounding box.}
	\vspace{-1.5mm}
	\resizebox{0.75\textwidth}{!}{
		\begin{tabular}{ccccccccc}
			\toprule
			\multirow{2}[4]{*}{\textbf{Dataset}} & \multirow{2}[4]{*}{\textbf{\#Images}} & \multirow{2}[4]{*}{\textbf{\#Categories}} & \multicolumn{3}{c}{\textbf{\#Instances}} & \multirow{2}[4]{*}{\textbf{Split}} & \multirow{2}[4]{*}{\textbf{Avg. Res. ($W\times H$)}} & \multirow{2}[4]{*}{\textbf{Year}} \\
			\cmidrule{4-6}          &     &     & \textbf{\textit{eS}}    & \textbf{\textit{rS}}    & \textbf{\textit{gS}}        &       &       &  \\
			\midrule
			TT100K \cite{117} & 8876  & 45 & 71    & 2800   & 6430     & train/test & $2048\times2048$ & 2016 \\
			EuroCity Persons \cite{115} & 32605 & 18  & 5318 & 28048 & 59190    & train/val & $1920\times1024$ & 2019 \\
			TJU-DHD Traffic \cite{144} & 50266  & 5 & 82 & 1189 & 20366     & train/val & $1624\times1200$ & 2021 \\
			SODA-10M \cite{145} & 10000 & 6 & 33 & 3061 & 10056    & train/val & $1920\times1080$ & 2021 \\
			\textbf{SODA-D} & 24828   & 9   & 25834     & 71064     & 102066         & train/val/test & $3407\times2470$     & 2022 \\
			\bottomrule
		\end{tabular}%
		\label{tab:Tab5-1}%
	}
\end{table*}%
\begin{table*}[h!]
	\scriptsize
	\centering
	% \caption{Comparisons of SODA-A and several related detection datasets under aerial scene.}
	\vspace{-1.5mm}
	\resizebox{0.75\textwidth}{!}{
		\begin{tabular}{ccccccccccc}
			\toprule
			\multirow{2}[4]{*}{\textbf{Dataset}} & \multirow{2}[4]{*}{\textbf{Annotation}} & \multirow{2}[4]{*}{\textbf{\#Images}} & \multirow{2}[4]{*}{\textbf{\#Categories}} & \multicolumn{3}{c}{\textbf{\#Instances}} & \multirow{2}[4]{*}{\textbf{Split}} & \multirow{2}[4]{*}{\textbf{Avg. Res.  ($W\times H$)}} & \multirow{2}[4]{*}{\textbf{Year}} \\
			\cmidrule{5-7}          &       &       &     & \textbf{\textit{eS}}    & \textbf{\textit{rS}}    & \textbf{\textit{gS}}        &       &       &  \\
			\midrule
			CARPK \cite{146} & HBB   & 1448  & 1 & 220    & 1716   & 1378     & train/test & $1280\times720$ & 2017 \\
			VisDrone \cite{147} & HBB   & 8629 & 10  & 78999 & 97251 & 108793 & train/val/test-dev & $1490\times957$ & 2021 \\
			AI-TOD \cite{118} & HBB   & 28036  & 8 & 193200 & 135566 & 17200 & train/val & $800\times800$ & 2021 \\
			DOTA \cite{30}  & OBB   & 2423 & 18  & 114045 & 94867 & 69934    & train/val & $2217\times2074$ & 2021 \\
			DIOR-R \cite{143} & OBB   & 23463 & 20 & 30938 & 37471 & 39697  & train/val/test & $800\times800$ & 2022 \\
			\textbf{SODA-A} & OBB   & 2513   & 9   & 304900    & 363738     & 168874         & train/val/test & $4761\times2777$     & 2022 \\
			\bottomrule
		\end{tabular}%
		\label{tab:Tab5-2}%
	}
	\vspace{-1em}
\end{table*}%

\begin{figure*}[t!]
	\centering
	\includegraphics[scale=0.55]{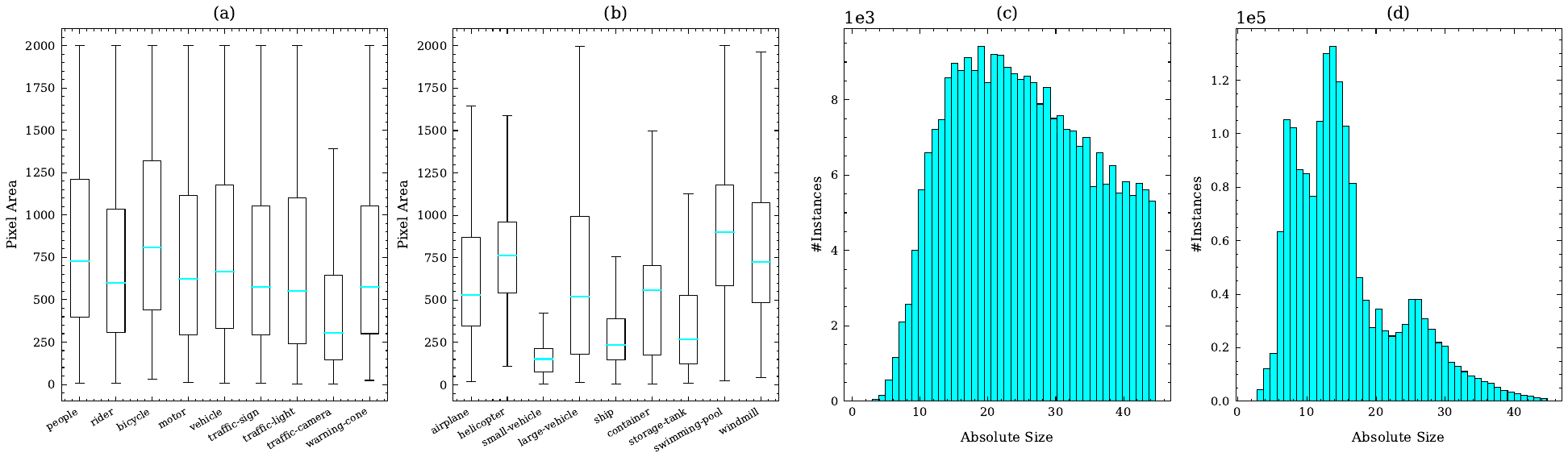} 
	\vspace{-1.5em}
	\caption{Category-wise area distribution of instances in SODA-D (a) and SODA-A (b), and overall scale distribution of instances in SODA-D (c) and SODA-A (d).}
	\label{fig:Fig5}
	\vspace{-1em}
\end{figure*}

\textbf{Dataset split}. Following the pioneering works \cite{6, 33}, we split the full image-set into three subsets: train-set, validation-set and test-set, and each subset occupies approximately $50\% :20\%: 30\%$ for SODA-D and $40\% :25\%: 35\%$ for SODA-A.

\textbf{Category selection}. Take the realistic value for applications and the intrinsic size into consideration, we select nine valuable categories for SODA-D: \textit{people}, \textit{rider}, \textit{bicycle}, \textit{motor}, \textit{vehicle}, \textit{traffic-sign}, \textit{traffic-light}, \textit{traffic-camera}, and \textit{warning-cone}. For SODA-A, we also annotate nine object classes: \textit{airplane}, \textit{helicopter}, \textit{small-vehicle}, \textit{large-vehicle}, \textit{ship}, \textit{container}, \textit{storage-tank}, \textit{swimming-pool}, and \textit{windmill}.

\textbf{Instance-level annotation}. The general principle to annotate SODA resembles that of general detection benchmarks \cite{6, 20, 30, 31, 32}, and the only difference lies in the \textit{ignore} regions. Enlightened by the previous works \cite{7, 8, 150}, we assign \textit{ignore} label to the two datasets when: 1) the instances belonging to the preset categories but with an area greater than 2000; 2) the objects that are excessively small and heavily occluded thus cannot be distinguished. In addition, we merge the \textit{ignore} regions as possible while avoiding surround valuable foreground instances. 

\subsection{Statistical Analysis}
We annotate 278433 instances for SODA-D and 872069 objects for SODA-A, and the number of instances for each category and that for three subsets are shown in Tab. \ref{tab:Tab4}. Also the example instances of each category are shown in Fig. \ref{fig:Fig4}. 

Next we highlight the most prominent feature of our dataset: \textbf{small size}. From Tab. \ref{tab:Tab5-1}, SODA-D and SODA-A both far exceed the existing mainstream object detection datasets under traffic and aerial scenarios on the amount of \textit{Small} objects, especially for \textit{extremely Small} ones. Moreover, we also show the category-wise area distribution and overall scale distribution of instances in SODA-D and SODA-A in Fig. \ref{fig:Fig5}. As can be seen from (a) and (b), the area of objects in our benchmarks falls into a relatively tight range (especially for \textit{traffic-camera} in SODA-D and \textit{small-vehicle} and \textit{ship} in SODA-A). Moreover, from (c) and (d) in Fig. \ref{fig:Fig5}, the size range of objects in SODA-D mainly comes to $\left[10, 30 \right]$ and for SODA-A, it is strikingly $\left[5, 15 \right]$. If we shed our light on the \textit{Small} objects, the average absolute size of SODA-D and SODA-A is $20.31$ pixels and $14.75$ pixels, respectively.

Except the small size and large volume, our SODA-D and SODA-A also exhibit several unique characters, as discussed next. 

\begin{figure*}[t!]
	\centering
	\includegraphics[scale=1.0]{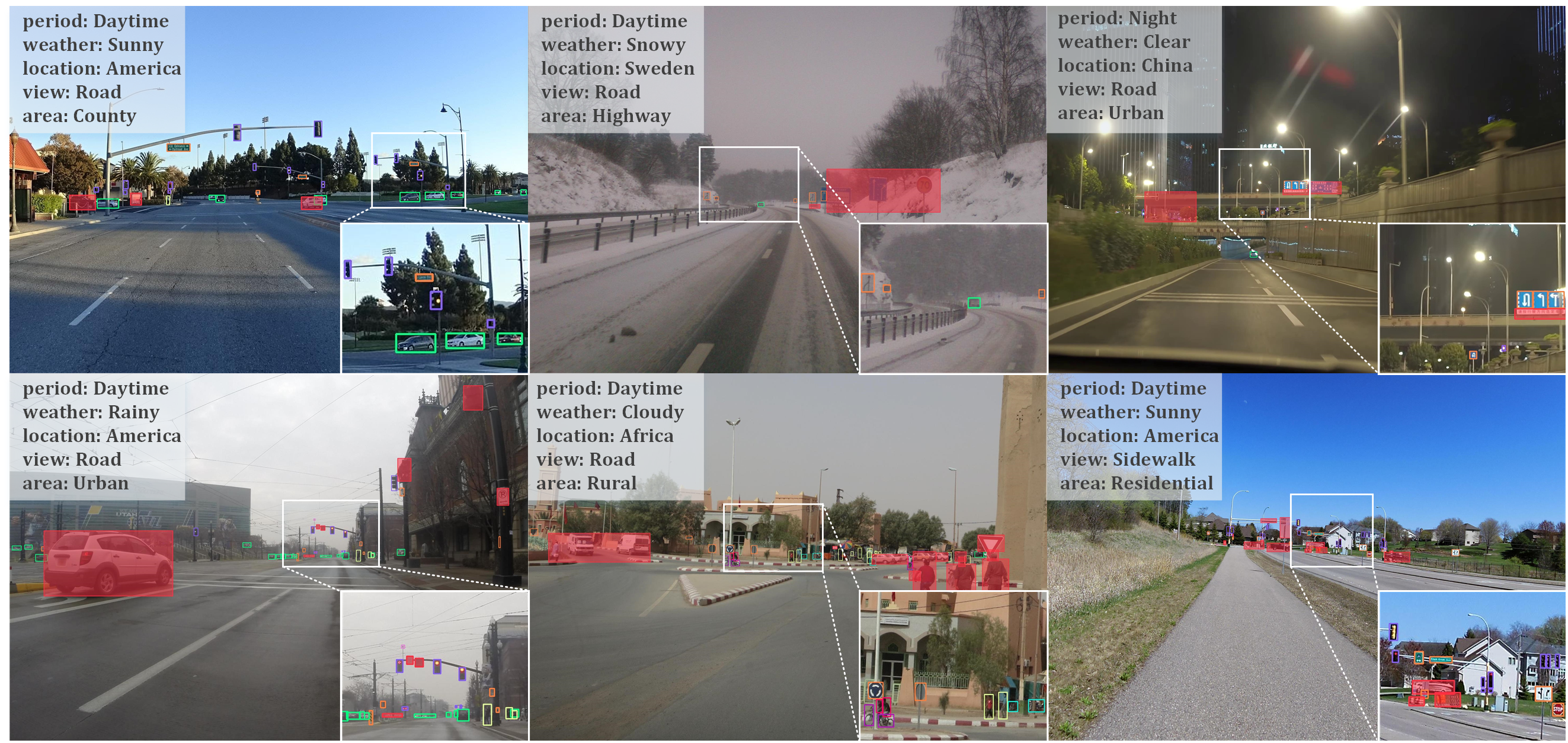} 
	\vspace{-1em}
	\caption{Example images under diversified conditions in our SODA-D dataset, where masked bounding boxes represent \textit{ignore} regions. Best viewed in zoom-in windows.}
	\label{fig:Fig6}
	\vspace{-1em}
\end{figure*}

\begin{figure}[t]
	\centering
	\vspace{-0.5em}
	\includegraphics[scale=0.485]{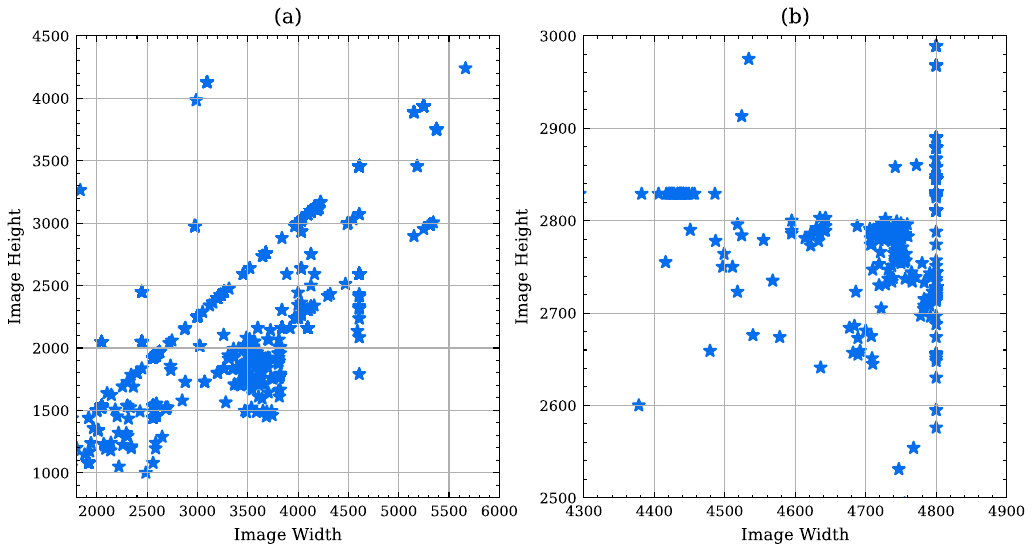} 
	\vspace{-1em}
	\caption{The distribution of image resolution in SODA-D (a) and SODA-A (b). Note that we randomly sample 2000 images to obtain the size profile for clear illustration.}
	\label{fig:Fig7}
	\vspace{-1.5em}
\end{figure}

\subsubsection{Data properties of SODA-D}
\textbf{Rich diversity}. Our SODA-D dataset inherits one of the most preeminent virtues of MVD: the rich diversity in terms of locations, weathers, period, shooting views and scenarios. Fig. \ref{fig:Fig6} shows some examples of our dataset covering various weather, view and illumination conditions. We believe that our diverse data could empower the model with the ability to generalize to different situations.

\textbf{High spatial resolution}. The images in SODA-D enjoy very high resolution and high quality, which is entailed for small or tiny object detection. In Fig. \ref{fig:Fig7}, we demonstrate the distribution of image resolution in SODA-D, and the average resolution at $3407\times 2470$ shows a clear predominance in comparison with previous datasets who focus on object detection under traffic scenes, as illustrated in Tab. \ref{tab:Tab5-1}.

\textbf{Ignore regions}. Our benchmark contains a mass of \textit{ignore} annotations (especially for SODA-D which has 153976 well-annotated \textit{ignore} regions), which is one of the most highlighted features. The \textit{ignore} definitions of \textit{Instance -level annotation} part in Sec. \ref{sec:Sec4.1} could maintain the stability of training and evaluation. Concretely, we deem that the prevailing detectors \cite{1, 3, 4, 5, 9, 45, 47, 50, 51, 52, 151} can handle the first situation well, hence it is not our concern. For the latter condition, our well-trained annotators are called for cautiously labeling the regions as ignore, when they cannot make confident judgment even at highest zoom-in level. And it will only bring error and instability if we insist on annotating these regions as foreground objects. To put it in another way, \textit{can we expect current algorithms to outperform human’s eyes}? Therefore, categorizing these regions into \textit{ignore} will not impose negative impact during evaluation process, and can guarantee the models concentrate on the authentic and valuable small objects.

\begin{figure*}[t!]
	\centering
	\includegraphics[scale=0.49]{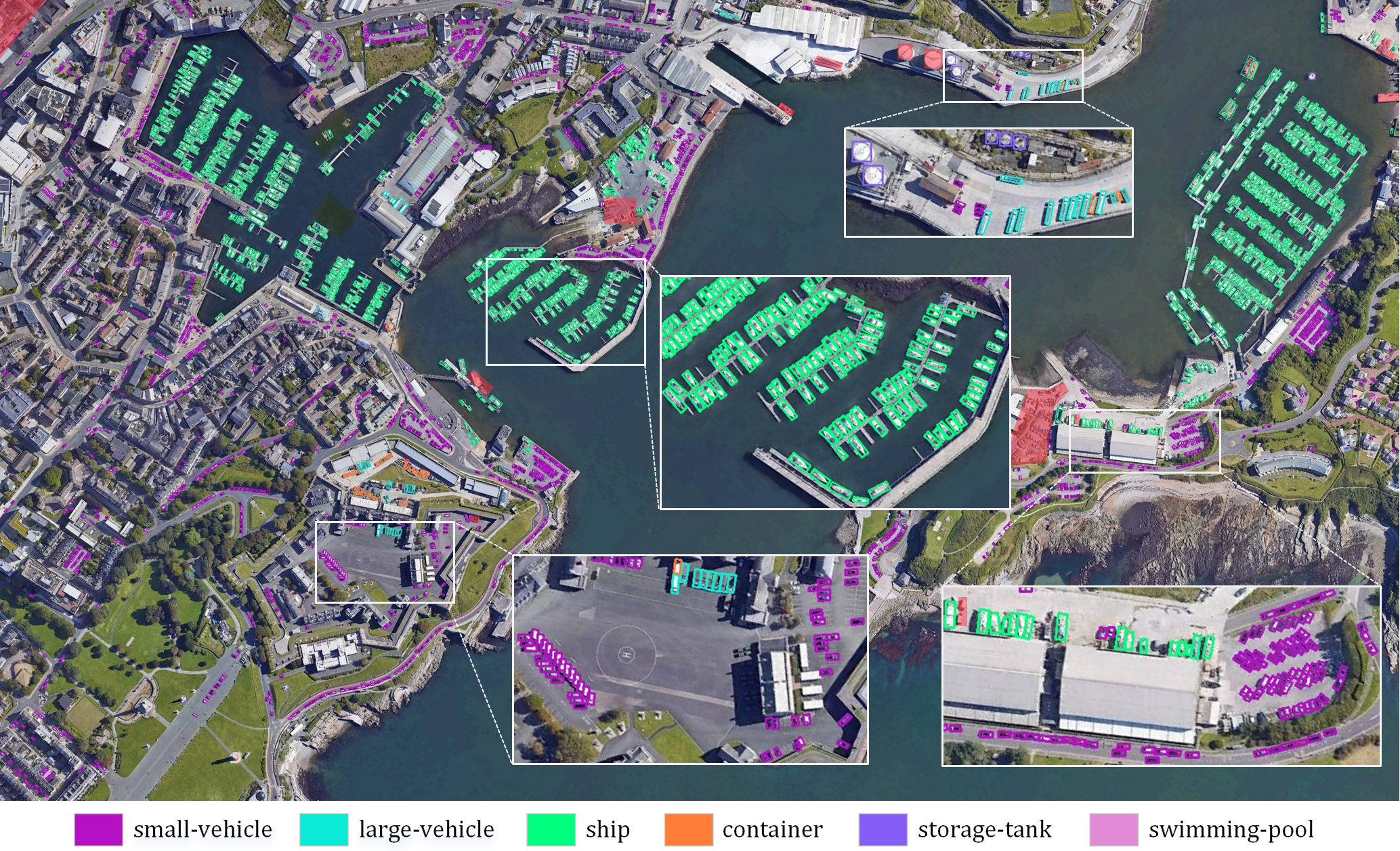} 
	\vspace{-1em}
	\caption{ An example image in SODA-A. The instances of different categories are best viewed in color and zoom-in windows, where masked areas denote the \textit{ignore} regions.}
	\label{fig:Fig8}
\vspace{-1.5em}
\end{figure*}

\subsubsection{Data properties of SODA-A}
We show an example image of SODA-A in Fig. \ref{fig:Fig8} and the local zoom-in windows exhibit the details of annotated instances.

\begin{figure}[t]
	\centering
	\includegraphics[scale=0.5]{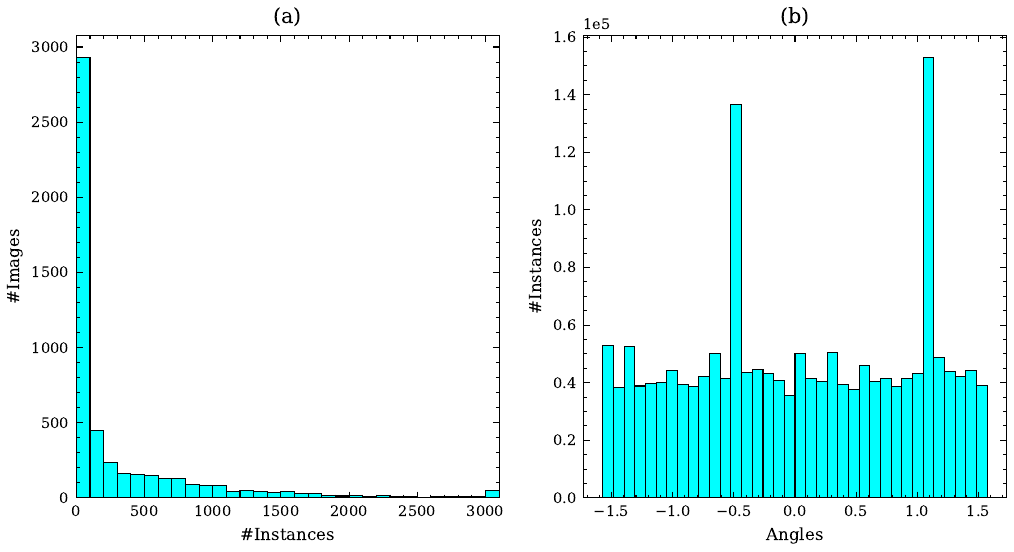}
	\vspace{-1em}
	\caption{Density distribution per image (a) and the orientation profile (b) of instances in SODA-A. Note that the number of images with more than 3000 instances were accumulated for clear demonstration of (a).}
	\label{fig:Fig9}
	\vspace{-1.5em}
\end{figure}

\textbf{Large density variation}. As demonstrated in Fig. \ref{fig:Fig9}, the number of instances per image in SODA-A varies significantly from 1 to 11134, which implies that our benchmark not only contains sparse condition but also includes numerous images where the objects positioned in extremely close proximity. Moreover, the average number of instances per image in SODA-A is 347.02, which is more than twice the number of DOTA (159.18). Such distribution literally calls for a robust model with the capacity of handling excessively clustered situation.

\textbf{Various orientations}. The instances in SODA-A can appear in an arbitrary-rotated fashion. We indicate the orientation distribution of SODA-A in Fig. \ref{fig:Fig9}, and the tilt angle of annotated instances distributes from ${{-\pi}/{2}}$ to ${{\pi}/{2}}$. Note that we do not follow the orientation definition in DOTA, because most objects with tiny size cannot convey sufficient visual cues to determine their head or tail.

\textbf{Diverse locations}. The images in SODA-A are collected from hundreds of cities around the world, which substantially enhances the data diversity in fact (\textit{e.g.}, the appearance of \textit{airplane} objects in our SODA-A can vary considerably). Furthermore, the concomitant intra-class variation and complicated background bring more challenges. 

\vspace{-1em}
\subsection{Comparisons with Previous Benchmarks}
Although there have been tremendous datasets for object detection, few of them dedicated to SOD task. Even so, we compare several related benchmarks with SODA to highlight its uniqueness.

\subsubsection{SODA-D}
\textbf{MVD}: Despite the SODA-D dataset is constructed on top of MVD, our intention is completely different from MVD. To be more specific, MVD concentrates on the pixel-level understanding of street scenes, while the proposed SODA-D highlights the detection of those objects with extremely small size under complicated driving scenarios.

\subsubsection{SODA-A}
\textbf{AI-TOD}: AI-TOD is built on several publicly available datasets, including DIOR \cite{20}, DOTA \cite{30}, VisDrone \cite{147}, xView \cite{155}, and Airbus-Ship$ \footnote[5]{\url{https://www.kaggle.com/c/airbus-ship-detection}}$. However, the above datasets were not designed for SOD task, hence more than 88$\%$ instances of AI-TOD come from the category \textit{vehicle}, leading to a non-negligible imbalance issue as shown in Fig. \ref{fig:Fig10}. Meanwhile, each category in our SODA-A contains adequate instances, except \textit{helicopter} class, and this advantage becomes more pronounced when considering the data volume (our SODA-A contains 837512 instances belonging to \textit{Small} object subset). In addition, the images in AI-TOD are cropped from existing datasets and the image resolution is fixed to $800\times 800$. More importantly, AI-TOD only provides horizontal annotations, which severely limits its capacity to approach objects accurately and to handle the densely-packed situation that is common and challenging for SOD in aerial images. In contrast, from Tab. \ref{tab:Tab5-1} and Fig. \ref{fig:Fig7}, our SODA-A possesses an average image resolution of $4761\times 2777$, and the well-annotated oriented boxes allow for large density cases and encourage more accurate localization.

\textbf{DOTA}: DOTA is the largest dataset for object detection in aerial images to date. Compared to DOTA, who puts emphasis on scale variation issue, we mainly focus on the small-scale objects which obstruct current detectors. Moreover, though DOTA contains substantial amounts of small objects, most of them centralized at \textit{small-vehicle}, as in Fig. \ref{fig:Fig10}.

\section{Experiments}
\subsection{Evaluation Protocol}
Following the evaluation protocols in COCO \cite{6}, we use the Average Precision (AP) to evaluate the performance of detectors. Concretely, as the paramount metric, the overall $AP$ is obtained by averaging the AP over 10 IoU thresholds between 0.5 and 0.95 (with an interval of 0.05) on \textit{Small} objects. $AP_{50}$ and $AP_{75}$ are computed at the single IoU thresholds of 0.5 and 0.75, respectively. Moreover, to highlight our concern for size-limited objects, the AP of four area subsets also are demonstrated, namely, $AP_{eS}$, $AP_{rS}$, $AP_{gS}$ and $AP_N$.

\begin{figure}[t]
	\vspace{-0.5em}
	\centering
	\includegraphics[scale=1.2]{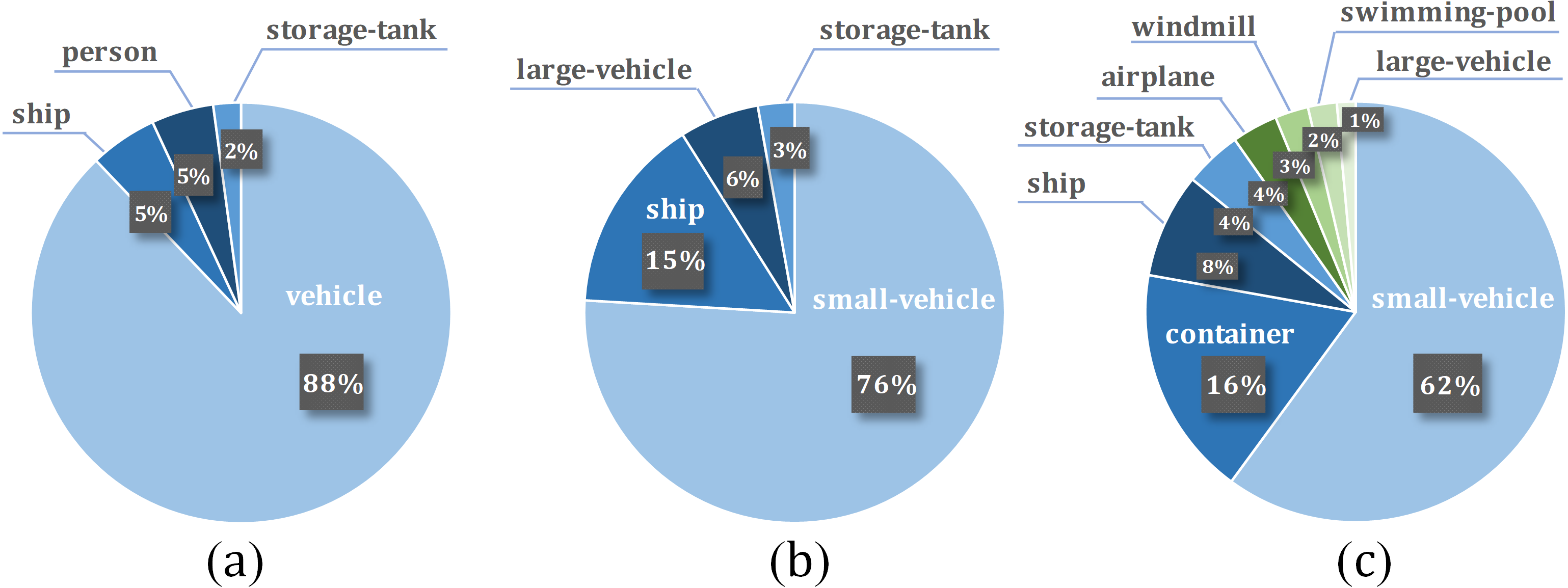} 
	\vspace{-1em}
	\caption{Class distribution of \textit{Small} instances in AI-TOD (a), DOTA (b) and SODA-A (c). Those categories with instances less than 2000 are not included.}
	\label{fig:Fig10}
	\vspace{-1em}
\end{figure}

\begin{table*}[t]
	\scriptsize
	\centering
	\caption{Baseline results on SODA-D test-set. All the models are trained with a ResNet-50 \cite{10} as the backbone except YOLOX (CSP-Darknet) \cite{YOLOX} and CornerNet (HourglassNet-104) \cite{51}. Schedule denotes the epoch setting during training, where '1$\times$' refers to 12 epochs and '50e' represents 50 epochs.}
	\vspace{-1em}
	\resizebox{\textwidth}{24mm}{
		\begin{tabular}{|c|c|c|ccc|cccc|c|c|}
			\hline
			Method & Publication & Schedule & $AP$  & $AP_{50}$  & $AP_{75}$  & $AP_{eS}$  & $AP_{rS}$  & $AP_{gS}$  & $AP_{N}$  & $\#$Param. & FLOPs \\
			\hline
			\hline
			Faster RCNN \cite{1} & TPAMI 2017 & 1$\times$    & 28.9  & 59.7  & 24.2  & 13.9  & 25.6  & 34.3  & 43.2  & 41.16M & 292.28G  \\
			Cascade RCNN \cite{151} & TPAMI 2021 & 1$\times$    & 31.2  & 59.9  & 27.8  & 14.1  & 27.5  & 37.1  & 46.9  & 68.95M & 320.07G  \\
			RetianNet \cite{3} & TPAMI 2020 & 1$\times$    & 28.2  & 57.6 & 23.7 & 11.9    & 25.2  & 34.1  & 44.2  & 35.68M & 299.50G  \\
			CornerNet \cite{51} & ECCV 2018 & 2$\times$    & 24.6 & 49.5 & 21.7 & 6.5 & 20.5 & 32.2 & 43.8  & 200.96M & 1104.06G  \\
			CenterNet \cite{47} & ArXiv 2019 & 70e    & 21.5  & 48.8  & 15.6  & 5.1  & 16.2  & 29.6  & 42.4  & 70.75M & 137.21G  \\
			FCOS \cite{4} & TPAMI 2022 & 1$\times$    & 23.9  & 49.5  & 19.9    & 6.9  & 19.4  & 30.9  & 40.9  & 31.86M & 284.53G  \\
			RepPoints \cite{152} & ICCV 2019 & 1$\times$    & 28.0  & 55.6  & 24.7  & 10.1 & 23.8  & 35.1  & 45.3  & 36.60M & 273.96G  \\
			ATSS \cite{153} & CVPR 2020 & 1$\times$    & 26.8 & 55.6 & 22.1 & 11.7 & 23.9 & 32.2 & 41.3  & 31.32M  & 290.79G \\
			Deformable-DETR \cite{52} & ICLR 2020 & 50e   & 19.2  & 44.8  & 13.7  & 6.3  & 15.4  & 24.9  & 34.2  & 35.17M & 739.11G  \\
			Sparse RCNN \cite{154} & CVPR 2021 & 1$\times$ &  24.2 & 50.3 & 20.3 & 8.8 & 20.4 & 30.2 & 39.4 & 105.96M  & 213.00G \\
			YOLOX \cite{YOLOX} & ArXiv 2021 & 70e & 26.7  & 53.4  & 23.0  & 13.6  & 25.1  & 30.9  & 30.4  & 8.94M & 48.11G  \\
			RFLA \cite{137} & ECCV 2022 & 1$\times$    & 29.7  & 60.2  & 25.2  & 13.2  & 26.9  & 35.4  & 44.6  & 41.16M & 292.06G  \\
			\hline
		\end{tabular}%
		\label{tab:Tab6}%
	}
	\vspace{-0.5em}
\end{table*}%

\begin{table*}[t]
	\scriptsize
	\centering
	\caption{Category-wise $AP$ of baseline detectors on SODA-D test-set.The training settings are consistent with Tab. \ref{tab:Tab6} and the full names of class abbreviation are as follows: t-sign (traffic-sign), t-light (traffic-light), t-camera (traffic-camera) and w-cone (warning-cone).}
	\vspace{-1em}
	\resizebox{\textwidth}{24mm}{
		\begin{tabular}{|c|ccccccccc|c|}
			\hline
			Method & \multicolumn{1}{c}{people} & \multicolumn{1}{c}{rider} & \multicolumn{1}{c}{bicycle} & \multicolumn{1}{c}{motor} & \multicolumn{1}{c}{vehicle} & \multicolumn{1}{c}{t-sign} & \multicolumn{1}{c}{t-light} & \multicolumn{1}{c}{t-camera} & \multicolumn{1}{c|}{w-cone} & $AP$ \\
			\hline
			\hline
			Faster RCNN \cite{1} & 35.8 & 16.5 & 12.5 & 23.1 & 44.1 & 45.8 & 37.8 & 14.3 & 30.5  & 28.9 \\
			Cascade RCNN \cite{151} & 39.2 & 18.0 & 14.5 & 24.2 & 47.4 & 48.1 & 39.8 & 15.2 & 33.4  & 31.2 \\
			RetianNet \cite{3} & 34.0 & 16.9 & 11.1 & 22.5 & 44.3 & 45.6 & 36.3 & 14.2 & 29.0 & 28.2 \\
			CornerNet \cite{51}  & 30.5 & 15.7 & 11.3 & 22.8 & 37.3 & 40.3 & 31.8 & 8.0 & 24.3 & 24.6\\
			CenterNet \cite{47}  & 25.6 & 12.8 & 9.5 & 19.9 & 32.9 & 35.8 & 27.6 & 9.3 & 20.4 & 21.5 \\
			FCOS \cite{4}  & 29.7 & 13.9 & 10.4 & 19.5 & 40.2 & 38.0 & 31.6 & 8.9 & 23.0 & 23.9 \\
			RepPoints \cite{152} & 36.0 & 15.9 & 10.8 & 21.6 & 44.8 & 45.6 & 37.3 & 12.7 & 27.7 & 28.0 \\
			ATSS \cite{153}  & 33.3 & 16.0 & 10.6 & 21.3 & 42.7 & 43.3 & 34.8 & 11.6 & 27.8 & 26.8 \\
			Deformable-DETR \cite{52} & 23.0 & 10.2 & 7.6 & 16.4 & 30.2 & 33.8 & 24.9 & 7.5 & 19.6 & 19.2 \\
			Sparse RCNN \cite{154} & 31.4 & 12.4 & 8.5 & 19.0 & 39.9 & 42.1 & 33.1 & 8.5 & 23.0 & 24.2 \\
			YOLOX \cite{YOLOX}  & 33.8 & 14.7 & 8.0 & 20.1 & 43.6 & 43.3 & 35.9 & 11.5 & 29.5 & 26.7 \\
			RFLA \cite{137}  & 37.1 & 19.1 & 13.4 & 24.2 & 45.0 & 45.8 & 37.5 & 14.7 & 30.5 & 29.7 \\
			\hline
		\end{tabular}%
		\label{tab:Tab7}%
	}
	\vspace{-0.5em}
\end{table*}%

\begin{table}[t]
	\centering
	\vspace{-1em}
	\caption{The $AP$ performance of baseline detectors with different backbone networks. All the models were trained for '$1\times$' schedule.}
	\vspace{-1em}
	\resizebox{\columnwidth}{14mm}{
		\begin{tabular}{|c|cccc|}
			\hline
			Method & ResNet-50 & ResNet-101 & Swin-T & ConvNext-T  \\
			\hline
			\hline
			Faster RCNN \cite{1} & 28.9 & 28.7 & 30.3 & 31.9 \\
			Cascade RCNN \cite{151} & 31.2 & 30.6 & 32.8 & 34.3 \\
			RetianNet \cite{3} & 28.2 & 27.8 & 28.8 & 29.7 \\
			FCOS \cite{4} & 23.9 & 24.3 & 29.2 & 29.1 \\
			RepPoints \cite{152} & 28.0 & 28.2 & 28.4 & 30.2 \\
			ATSS \cite{153} & 26.8 & 26.7 & 27.2 & 27.9 \\
			Sparse RCNN \cite{154} & 24.2 & 25.3 & 24.9 & 25.2 \\
			\hline
		\end{tabular}%
		\label{tab:Tab8}%
	}
	\vspace{-1em}
\end{table}%

\subsection{Implementation Details}
To conduct fair comparisons of several benchmarking baselines, all the experiments on SODA-D and SODA-A are implemented on top of the open source object detection toolbox mmdetection$ \footnote[6]{\url{https://github.com/open-mmlab/mmdetection}}$\cite{156} and mmrotate$ \footnote[7]{\url{https://github.com/open-mmlab/mmrotate}}$\cite{157}, respectively. Directly feeding the high-resolution images in SODA to deep model is infeasible due to the GPU memory limitation, hence we crop original images into a series of $800\times 800$ patches with a stride of 650. These patches will be resized to $1200\times 1200$ during training and testing, which could partly alleviate the information loss caused in the feature extraction stage. Noting the patch-wise detection results will be first mapped to the original images, on which Non Maximum Suppression (NMS) was performed to prune out redundant predictions. We use 4 NVIDIA GeForce RTX 3090 GPUs to train the models, and the batch size is set to 8 for the experiments of SODA-D and 4 for that of SODA-A, where the angle ranges is $\left[ -\pi/{2}, \pi/2\right)$. Only random flip was used for augmentation during training, and more details and hyperparamter settings please refer to Sec. \ref{app:sodad details} and Sec. \ref{app:sodaa details} in Appendix.

\begin{figure*}[t!]
	\centering
	\includegraphics[scale=1.1]{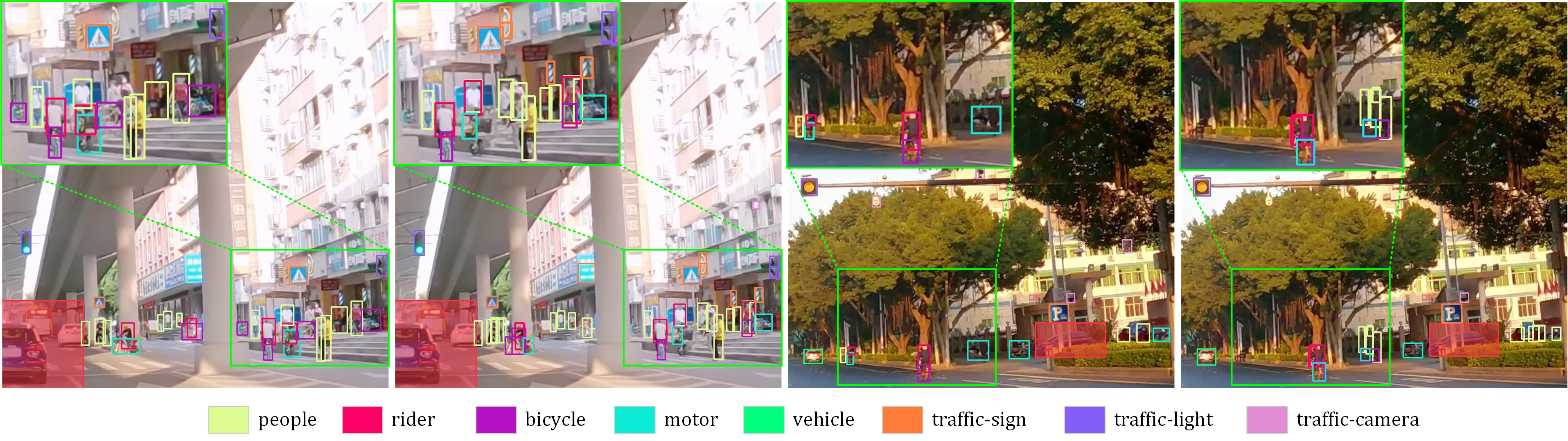} 
	\vspace{-1em}
	\caption{Qualitative results of Cascade RCNN \cite{151} on SODA-D test-set. Columns 1 and 3 denote the ground-truth annotations and columns 2 and 4 stand for the predictions. Best viewed in color and zoom-in windows, where masked bounding boxes represent \textit{ignore} regions. Only predictions with confidence scores larger than 0.3 are demonstrated.}
	\label{fig:Fig11}
	\vspace{-1em}
\end{figure*}

\subsection{Results Analysis on SODA-D}
In this section, we perform a rigorous evaluation of several representative methods on our SODA-D dataset, and provide in-depth analyses on top of the results. Moreover, we conduct several experiments to investigate the effect of label assignment and loss designs to SOD, details can be found in Sec. \ref{app:label assignmen} and Sec. \ref{app:loss function} in Appendix.

\subsubsection{Benchmarking Results}\label{sec:Sec5.3.1}
Tab. \ref{tab:Tab6} reports the results of 12 representative methods on SODA-D test-set. From the table, we can find that Faster RCNN \cite{1} scores $28.9\%$ on $AP$, and benefiting from the cascade structure, Cascade RCNN \cite{151} attains the best performance with an $AP$ of $31.2\%$ and an impressive $AP_{75}$ of $27.8\%$, which steadily outperform other detectors. On top of Faster RCNN, RFLA \cite{137} achieves $29.7\%$ $AP$, though meanwhile, the $AP_{eS}$ actually drops 0.7 points, showing that the devised assignment might not be suitable for those instances with excessively limited sizes. One-stage detector RetinaNet \cite{3} scores $28.2\%$ $AP$ which is close to Faster RCNN, but there exists a huge gap ($11.9\%$ \textit{v.s.} $13.9\%$) when comes to the $AP_{eS}$, and when the object size gets larger, such difference becomes smaller, which reveals that the misalignment issue imposes a significantly severe impact on tiny objects. Similarly, though RepPoints \cite{152} can obtain an overall $AP$ of $28.0\%$, but the $AP_{eS}$ metric ($10.1\%$) is largely behind Faster RCNN and RetinaNet. This phenomenon indicates that point representation, in comparison to its box counterpart, may not be a good choice for small objects, but shows great potential for large ones. For anchor-free detectors, ATSS \cite{153} can achieve $26.8\%$ $AP$ on our SODA-D test-set, which is superior to FCOS \cite{4} ($23.9\%$), and the latter behaves badly on \textit{extremely Small} objects ($6.9\%$). This may partly originates from the occlusion challenge of our dataset, also known as the ambiguous sample problem. CenterNet \cite{47} and CornerNet \cite{51} only obtain an $AP$ of $21.5\%$ and $24.6\%$, respectively. It can be noticed that even with more training epochs, the performances of CenterNet and CornerNet are remarkably inferior to that of anchor-based methods, and the disparity becomes more staggering for \textit{extremely Small} and \textit{relatively Small} objects. YOLOX \cite{YOLOX} can obtain competitive results ($26.7\% AP$ and $13.6 \% AP_{eS}$) when compared to other anchor-free counterparts though meanwhile struggles on the objects of large areas. For the query-based detector, Sparse RCNN \cite{154} achieves $24.2 \%$ $AP$ which is comparable to FCOS. Though exploiting multi-scale deformable attention to reduce high computation in encoder and enabling the access of high-resolution features, Deformable DETR \cite{52} only delivers $19.2\%$ $AP$, lagging noticeably behind other competitors even with more training epochs. This performance gap may reveal that the sparse query paradigm could not cover small objects adequately.

% tie for the second place in ranking

\subsubsection{Category-Wise Results}
We also list the category-wise results on Tab. \ref{tab:Tab7}, from which we can see that the AP of \textit{rider}, \textit{bicycle}, \textit{motor} and \textit{traffic-camera} are clearly inferior to other categories, we deem that the root cause of this phenomenon comes from two-fold. 1) Class-imbalance issue. These categories contain less samples compared to other classes, \textit{e.g.}, only 2560 samples included in \textit{bicycle} category. 2) The limited area. For instance, nearly half of the \textit{traffic-camera} objects possess an area less than 256 pixels, as demonstrated in Fig. \ref{fig:Fig10}. In other words, this phenomenon corroborates previous findings, \textit{i.e.}, the detection difficulty increases sharply when the object size gets smaller. 

\subsubsection{Baseline Detectors with Different Backbones}
Tab. \ref{tab:Tab8} shows the performance of baseline detectors with different backbone networks. Compared to ResNet-50, ResNet-101 only brings a slight improvement even degrades the performance (see Cascade RCNN and RetinaNet). This phenomenon substantiates previous hypothesis that deeper models might not be better for the size-limited objects and moreover, the highly structural representations in deeper layers which hardly contain small object cues are suboptimal for detection. Swin-T \cite{158} yields substantial improvements for all detectors, especially for FCOS (+5.3 points). This impressive performance reveals the powerful representation ability of shifted-window scheme for small objects, and could shed more light on the subsequent feature extractor design of SOD. Not surprisingly, most detectors with ConvNext-T \cite{CN-T} as the backbone achieve the best performance, exhibiting good robustness and potential in capturing the finer representations of small objects.

\subsubsection{Qualitative Results}
Fig. \ref{fig:Fig11} demonstrates the visualization results of Cascade RCNN on SODA-D test-set. The first pair shows the challenge under complicated background and heavy occlusion, where the detector can hardly learn discriminative representation from instances with limited sizes and is inclined to lose those instances resembled the background. In addition, identifying those partly occluded objects is even more challenging. The second pair represents the detections of low illumination, in which the detector fails to recognize those instances under the shadow, still less predicts accurate bounding boxes. More qualitative results are exhibited in the Supplementary material.

\vspace{-1em}
\subsection{Results Analysis on SODA-A}
Based on SODA-A, we investigate the performance of several leading methods of oriented object detection. Also, considering our SODA-A contains densely packed issue, we explore the impact of proposal number for the final performance, please refer to Sec. \ref{app:proposals} of Appendix.

\begin{table*}[t]
	\centering
	\caption{Baseline results on SODA-A test-set. All the models are trained with a ResNet-50 as the backbone. Schedule denotes the epoch setting during training, where '1$\times$' refers to 12 epochs.}
	\vspace{-1em}
	\resizebox{\textwidth}{18mm}{
		\begin{tabular}{|c|c|c|ccc|cccc|c|c|}
			\hline
			Method & Publication & Schedule & $AP$  & $AP_{50}$  & $AP_{75}$  & $AP_{eS}$  & $AP_{rS}$  & $AP_{gS}$  & $AP_{N}$  & $\#$Param. & FLOPs \\
			\hline
			\hline
			Rotated Faster RCNN \cite{1} & TPAMI 2017 & 1$\times$    & 32.5 & 70.1  & 24.3  & 11.9  & 27.3  & 42.2 & 34.4 & 41.14M & 292.25G \\
			Rotated RetinaNet \cite{3} & TPAMI 2020 & 1$\times$    & 26.8  & 63.4  & 16.2  & 9.1  & 22.0  & 35.4  & 28.2 & 36.16M & 800.21G \\
			RoI Transformer \cite{159} & CVPR 2019 & 1$\times$    & 36.0  & 73.0  & 30.1  & 13.5  & 30.3  & 46.1 & 39.5  & 55.08M & 306.20G \\
			Gliding Vertex \cite{160} & TPAMI 2021 & 1$\times$    & 31.7 & 70.8 & 22.6 & 11.7 & 27.0 & 41.1 & 33.8 & 41.14M & 292.25G \\
			Oriented RCNN \cite{161} & ICCV 2021 & 1$\times$    & 34.4  & 70.7  & 28.6  & 12.5  & 28.6  & 44.5 & 36.7 & 41.13M & 292.44G \\
			S$^2$A-Net \cite{162} & TGRS 2022 & 1$\times$    & 28.3  & 69.6  & 13.1  & 10.2  & 22.8  & 35.8 & 29.5 & 38.64M & 732.74G \\
			DODet \cite{163} & TGRS 2022 & 1$\times$  & 31.6 & 68.1 & 23.4 & 11.3 & 26.3 & 41.0 & 33.5 & 69.34M & 555.49G \\
			Oriented RepPoints \cite{ORep} & CVPR 2022 & 1$\times$  & 26.3  & 58.8  & 19.0  & 9.4  & 22.6  & 32.4  & 28.5 & 55.66M & 827.21G \\
			DHRec \cite{DHRec} & TPAMI 2022 & 1$\times$  & 30.1 & 68.8 & 19.8 & 10.6 & 24.6 & 40.3 & 34.6 & 31.99M & 792.76G \\
			\hline
		\end{tabular}%
		\label{tab:Tab9}%
	}
	\vspace{-0.5em}
\end{table*}%

\begin{table*}[t]
	\centering
	\caption{Category-wise $AP$ of baseline detectors on SODA-A test-set. The training settings are consistent with Tab. \ref{tab:Tab9} and the full names of class abbreviation are as follows: s-vehicle (small-vehicle), l-vehicle (large-vehicle), s-tank (storage-tank) and s-pool (swimming-pool).}
	\vspace{-1em}
	\resizebox{\textwidth}{16.5mm}{
		\begin{tabular}{|c|ccccccccc|c|}
			\hline
			Method & airplane & helicopter & s-vehicle & l-vehicle & ship  & container & s-tank & s-pool & windmill & $AP$ \\
			\hline
			\hline
			Rotated Faster RCNN \cite{1} & 49.4 & 18.1 & 33.4 & 19.6 & 43.5 & 29.8 & 42.8 & 34.1 & 21.9 & 32.5 \\
			Rotated RetinaNet \cite{3} & 42.0 & 16.8 & 29.9 & 10.0 & 35.1 & 23.7 & 35.1 & 30.7 & 18.1 & 26.8 \\
			RoI Transformer \cite{159} & 53.2 & 21.4 & 36.1 & 25.9 & 46.4 & 35.7 & 44.6 & 36.9 & 23.5 & 36.0 \\
			Gliding Vertex \cite{160} & 46.7 & 12.8 & 33.3 & 21.9 & 43.4 & 29.8 & 43.3 & 31.2 & 22.7 & 31.7 \\
			Oriented RCNN \cite{161} & 52.2 & 20.2 & 34.4 & 24.4 & 45.2 & 32.1 & 43.1 & 36.3 & 22.2  & 34.4 \\
			S$^2$A-Net \cite{162} & 41.5 & 20.4 & 31.2 & 14.0 & 36.7 & 26.1 & 29.6 & 33.8 & 21.6 & 28.3 \\
			DODet \cite{163} & 49.4 & 19.8 & 32.1 & 17.3 & 41.3 & 26.0 & 42.2 & 34.7 & 21.3 & 31.6 \\
			Oriented RepPoints \cite{ORep} & 51.7 & 8.5 & 30.3 & 2.6 & 28.0 & 19.6 & 40.3 & 33.2 & 21.9 & 26.3 \\
			DHRec \cite{DHRec} & 45.5 & 17.2 & 31.0 & 15.6 & 38.5 & 28.5 & 38.8 & 34.5 & 20.9 & 30.1 \\
			\hline
		\end{tabular}%
		\label{tab:Tab10}%
	}
	\vspace{-0.5em}
\end{table*}%

\begin{table}[t]
	\centering
	\vspace{-1em}
	\caption{The $AP$ performance of baseline detectors on SODA-A test-set with different backbone networks. All the models were trained for '1$\times$' schedule.}
	\vspace{-1em}
	\resizebox{\columnwidth}{12mm}{
		\begin{tabular}{|c|cccc|}
			\hline
			Method & ResNet-50 & ResNet-101 & Swin-T & ConvNext-T \\
			\hline
			\hline
			Rotated Faster RCNN \cite{1} & 32.5 & 32.7 & 33.6 & 34.3 \\
			Rotated RetinaNet \cite{3} & 26.8  & 26.8 & 23.3 & 21.7 \\
			RoI Transformer \cite{159} & 36.0 & 35.8 & 36.1 & 37.5 \\
			Gliding Vertex \cite{160} & 31.7 & 32.0 & 32.9 & 34.0 \\
			Oriented RCNN \cite{161} & 34.4 & 34.4 & 35.1 & 35.9 \\
			S$^2$A-Net \cite{162} & 28.3 & 28.3 & 26.0 & / \\
			Oriented RepPoints \cite{ORep} & 26.3 & 26.7 & 26.2 & 25.7 \\
			\hline
	\end{tabular}%
		\label{tab:Tab11}%
	}
	\vspace{-2em}
\end{table}%

\subsubsection{Benchmarking Results}
Tab. \ref{tab:Tab9} shows the results of nine representative methods on SODA-A test-set. RoI Transformer \cite{159} achieves top performance with $36.0\%$ $AP$. This remarkable success can be attributed to its powerful proposal generator, in which rotated proposals produced by the RRoI Learner can guarantee the high recall of small objects. By revising vanilla Faster RCNN to output an additional angle prediction, Rotated Faster RCNN \cite{1} scores $32.5\%$ on $AP$, which validates the robustness of this prevailing method again. Oriented RCNN \cite{161} obtains a relatively high performance both at overall $AP$ ($34.4\%$). Thanks to its efficient oriented RPN, Oriented RCNN can generate high-quality proposals with negligible parameter grow. From the results of RoI Transformer and Oriented RCNN, we can see that high-quality proposals are of great significance to small object detection, particularly for the densely packed objects. Gliding Vertex \cite{160} and DODet \cite{163} both resort to novel representations for oriented objects, the former learns four gliding offsets to corresponding sides while the latter utilizes aspect ratio and area to denote an object. Gliding Vertex achieves $31.7 \%$ $AP$ which is comparable to DODet ($31.6 \%$). For one-stage detectors, Rotated RetinaNet \cite{3} achieves $26.8\%$ $AP$ and lags largely behind two-stage ones. This is because SODA-A contains considerable excessively small objects that one-stage paradigm cannot handle well, as discussed in Sec. \ref{sec:Sec5.3.1}. S$^2$A-Net \cite{162} designs feature alignment module to alleviate the misalignment problem, and finally achieves an $AP$ with $28.3\%$. Though it can substantially increase the score of $AP_{50}$, the concomitant performance decline on the $AP_{75}$  metric can be non-negligible (-3.3 points) when compared to Rotated RetinaNet, which indicates that the performance gain of S$^2$A-Net is likely to come at the cost of subsequent regression accuracy. Oriented RepPoints \cite{ORep} achieves $26.3\%$ points on $AP$ metric which is slightly inferior to Rotated RetinaNet, exhibiting such point set representation is unamiable for small objects in aerial scenario, especially for those with large aspect ratios which will be discussed in next section. By exploiting two horizontal rectangles to encode the multi-oriented object, DHRec \cite{DHRec} disposes the discontinuity problem subtly and achieves $30.1\%$ $AP$ which is significantly superior to its one-stage counterparts with least parameters.

\subsubsection{Category-Wise Results}
Category-wise results of baseline algorithms on SODA-A test-set are shown in Tab. \ref{tab:Tab10}. The AP of \textit{helicopter} category is observably below that of other classes due to limited instance numbers. The objects of \textit{large-vehicle} and \textit{container} with elongated structure challenge the regression branch especially for Oriented RepPoints, and moreover, Gliding Vertex and DODet have comparable results yet perform variably on different categories, which can be attributed to the different representation about oriented objects.

\begin{figure*}[t]
	\centering
	\includegraphics[scale=1.1]{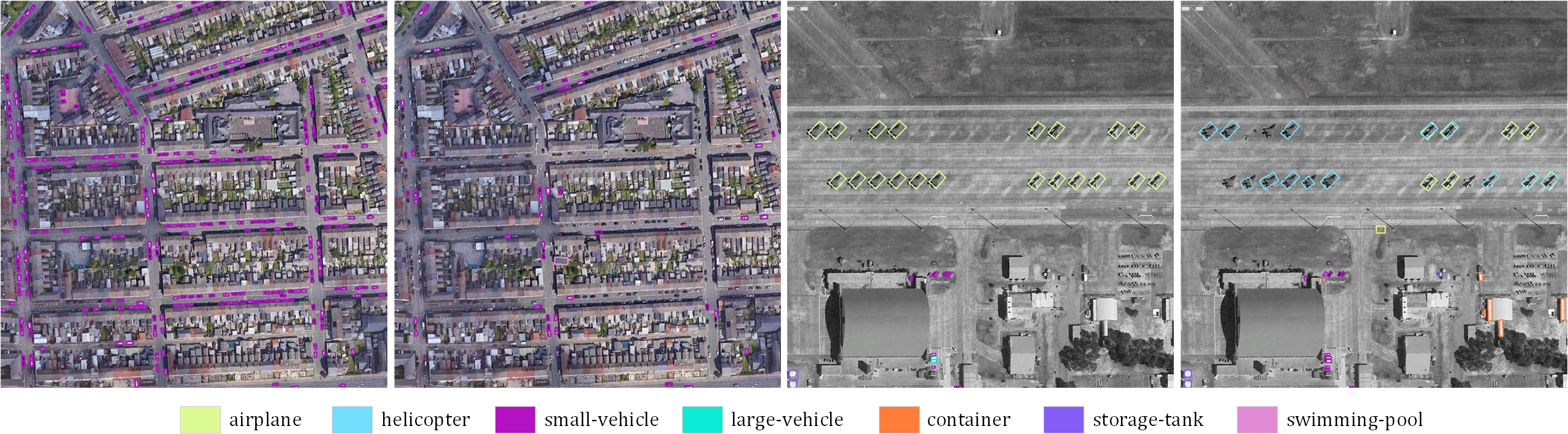} 
	\vspace{-1em}
	\caption{Qualitative results of Oriented RCNN \cite{161} on SODA-A test-set. Columns 1 and 3 represent the ground-truth annotations and columns 2 and 4 denote the predictions. Best viewed in color. Only predictions with confidence scores larger than 0.3 are demonstrated.}
	\label{fig:Fig12}
	\vspace{-1em}
\end{figure*}

\vspace{-1em}
\subsubsection{Baseline Detectors with Different Backbones}
Tab. \ref{tab:Tab11} shows the performance of baseline detectors with different backbone networks. Similar to the results on SODA-D, we can see that ResNet-101 only brings slight performance improvement even decline. 
However, when Swin-T backbone was employed to extract the features, two fundamentally distinct phenomena occur simultaneously. For RPN-based detectors, Swin-T can yield varying levels of performance gain (from 0.1 points to 1.2 points), but for RPN-free detectors, Swin-T causes substantial performance decline (-3.5 points for Rotated RetinaNet and -2.3 points for S$^2$A-Net), which is completely different from the results on SODA-D. We conjecture this disparity lies in the limited ability of Swin-T to cope with dense distribution when the detector suffers from misalignment issue, particularly for those objects with extremely close proximity. 
When taking ConvNext-T as the backbone network the general trend is similar to Swin-T, those RPN-free detectors suffer from more severe misalignment issue because there exists a huge gap between the object regions and horizontal priors.

\vspace{-1em}
\subsubsection{Qualitative Results}
We visualize the detection results of Oriented RCNN on SODA-D test-set in Fig. \ref{fig:Fig12}. The first pair shows the results of tiny instances and only very few of them were detected, demonstrating that detecting tiny objects is a massive challenge for current detectors, even with top performance. The second pair exhibits the detections of low contrast, of which \textit{airplane} instances possess similar visual feature with background and the model confuses them with \textit{helicopter}. Moreover, because the detailed information which is conducive for identification is hardly retained, the model is likely to utilize visual appearance for recognition instead, which unavoidably results in false positives and incorrect predictions (see the \textit{container} predictions). More qualitative results are exhibited in the Supplementary material.

\section{Conclusion and Outlook}
We presented a systematic study on small object detection. Concretely, we exhaustively reviewed hundreds of literature for SOD from the perspective of algorithms and datasets. Moreover, to catalyze the progress of SOD, we constructed two large-scale benchmarks under driving scenario and aerial scene, dubbed SODA-D and SODA-A. SODA-D comprises 278433 instances annotated with horizontal boxes, while SODA-A includes 872069 objects with oriented boxes. The well-annotated datasets, to the best of our knowledge, are the first attempt to large-scale benchmarks tailored for small object detection, and could serve as an impartial platform for benchmarking various SOD methods. On top of SODA, we performed a thorough evaluation and comparison of several representative algorithms. Based on the results, we discuss several potential solutions and directions for future development of SOD task.

\textbf{Effective feature extractor for small objects}. As alluded to in the results, deeper backbone networks might not be conducive to extract high-quality feature representations for small objects. Designing an effective backbone, which enjoys powerful feature extraction capability while avoiding high computational cost and information loss, is of paramount importance. 

\textbf{High-quality hierarchical representation}. FPN is an indispensable part in small object detection. Nevertheless, current feature pyramid architecture is suboptimal for SOD, owing to the heuristic pyramid level assignment strategy, few samples were assigned to higher levels (actually only $P_2$ feature is responsible to the detection during our benchmark experiments). Consequently, the high-level layers are optimized in an implicit and indirect manner which may hamper the fusion quality. Moreover, detecting on low-level feature maps brings heavy computational burden. Thus, an efficient hierarchical feature architecture tailored for SOD task is in high demand.

\textbf{Optimized label assignment strategy}. As we discussed in Sec. \ref{sec:Sec2.3.1} and Sec. \textcolor{green}{B.2} of Appendix, albeit the current label assignment schemes perform well on generic object detection and large objects, they still struggle on the instances of extremely small sizes, neither the overlap-based strategies nor the distribution-based ones. Therefore, designing an optimized strategy to assign sufficient positive samples for size-limited instances can substantially stabilize the training procedure and boost the performance further.

\textbf{Proper evaluation metric for SOD}. The multiple IoU thresholds-based evaluation process has been the de facto standard for validating the effectiveness of methods in generic object detection. However, such ubiquitous metric is too stringent for those instances with extremely sizes. In other words, the top priority of small object detection under some specific scenarios is to recognize the objects and obtain their rough locations instead of obsessing how accurate they are. Hence, it is impractical to pursue precise detections of small objects when the model cannot find them. Consequently, borrowing the experience of other fields such as crowd counting and devising a proper metric to guide the training and inference of SOD architectures under some specific scenes plays a significant role in future development.

\vspace{-1em}
\ifCLASSOPTIONcompsoc
  \section*{Acknowledgments}
\else
  \section*{Acknowledgment}
\fi
%\vspace{-0.5em}
We thank Peter Kontschieder for the constructive discussions and feedback, as well as their high-quality Mapillary Vistas Dataset.

\ifCLASSOPTIONcaptionsoff
  \newpage
\fi



%\bibliographystyle{IEEEtran}
%\begin{thebibliography}{1}
%\end{thebibliography}

\bibliographystyle{IEEEtran}
\bibliography{SODA}

\clearpage

\begin{appendices}
	\setcounter{table}{0}
	\setcounter{figure}{0}
	\renewcommand{\thetable}{\Alph{section}\arabic{table}}
	\renewcommand\thefigure{\Alph{section}\arabic{figure}}

	\section{Benchmarks}\label{app:bench}
	In this section, we first state the major considerations about why we choose the driving and aerial scenarios to construct our benchmark, and then the details about data cleaning and instance-level annotation are demonstrated.
	
	\subsection{Scene Selection}
	To acquire a vast collection of small instances for training robust deep models, we carefully choose the driving scene and aerial scene to construct our datasets. Our prime motivations come as follow:
	
	1. To simulate the real environment and capture size-limited objects, we need to increase the shooting distance, even so, the objects in the aforementioned two scenarios can still be identified due to their natural sizes and distributions. Moreover, the instances with similar sizes often occur intensively, this promises that we can obtain sufficient and valuable small instances.
	
	2. The annotation types adopted in the two benchmarks are Horizontal Bounding Box (HBB) and Oriented Bounding Box (OBB), which actually correspond to two of the most fundamental detection tasks, horizontal object detection and oriented object detection. That is to say, our benchmarks are amenable for most of the SOD algorithms to conduct the evaluation and comparison.
	
	3. These two scenes are both in high demand for SOD task: autonomous driving requires decision making based on the reliable and real-time understanding of complicated surroundings, where the objects far from the vehicle or occluded by other instances are with limited sizes, posing great challenges to the perceptron system. Meanwhile, overhead-view image analysis has an urgency to handle small objects in terms of the large flying altitude and various shooting views.
	
	\subsection{Data Cleaning}
	For SODA-D, the images that are visibly affected by artifacts, lens flare, strong motion blur and other factors which impede the subsequent annotation process were removed. Moreover, duplicated images collected from different websites were cleaned either. For SODA-A, we eliminate those images with noticeable blur and artifacts.
	
	\subsection{License Declaration}
	Our two benchmarks are freely available under the CC-BY-SA license agreement $ \footnote[1]{\url{https://creativecommons.org/licenses/by-nc/4.0/}}$.
	
	\subsection{Data Annotation}
	\textbf{Annotation tools}. As we alluded to in the text, the annotation type for the two benchmarks is different. Specifically, we annotate the objects of SODA-D with horizontal bounding boxes, and the instances of SODA-A are annotated with polygons which is in line with the pioneering works \cite{20, 30}. To precisely and efficiently annotate the instances with limited sizes in our database, we use Labelimg $ \footnote[2]{\url{https://github.com/tzutalin/labelImg}}$ and Labelme $ \footnote[3]{\url{https://github.com/wkentaro/labelme}}$ toolkits to conduct the annotations of SODA-D and SODA-A, which both allow for high-degree zoom-in operation, enabling the fine-grained annotations.
	
	\textbf{Instance-level annotation.}
	The annotation procedure is consistent with the general detection benchmarks \cite{6, 20, 30, 31, 32}. Concretely, for SODA-D, the annotators need to find the instances belonging to the predefined categories and then just draw tight bounding box enclosing the targets. Hence, here we put the emphasis on describing our annotation of SODA-A.
	
\begin{figure}[h]
		\centering
		\includegraphics[scale=1.25]{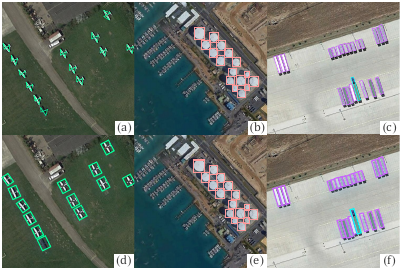} 
		\vspace{-1em}
		\caption{Three types of annotations used in SODA-A, \textit{i.e.}, crisscross polygon (a), horizontal bounding box (b) and enclosed polygon (c). The bottom (d, e, f) are the visualization results of oriented bounding boxes converted from original annotations.}
		\label{fig:FigA1}
	\end{figure}

	To efficiently perform the labeling process, we tailor optimal annotation strategies for different categories. For \textit{airplane} and \textit{helicopter} category, we design a new type of annotation method, crisscross annotation, which only requires four extreme points and is more appropriate to the objects with cruciform structures. In addition, we simply adopt horizontal box for \textit{storage-tank} and \textit{windmill} category. For remaining classes, the annotators use Labelme toolkit to create enclosed polygons along the contours of instances. Finally, the post-processing code was employed to convert
	the above annotations to unified oriented bounding box annotations. The visualization of the three types of annotations and converted oriented bounding boxes are shown in Fig. \ref{fig:FigA1}.
	
	%The preliminary annotation stage of the two benchmarks costs about 6150 hours in total, and involves 16 well-trained annotators.  After that, we install a two-step quality assurance procedure to guarantee the overall annotation quality.
	%
	%\textbf{Quality assurance}. As stated in \cite{148, 149}, the annotation quality of a dataset is the dominating factor for its lifespan as a benchmark, hence we install a two-step quality assurance procedure after the initial annotation. Concretely, at first step, each of the 16 professional annotators is responsible for reviewing the annotations of others and reporting back the existing problems to the original annotator to minimize missing or incorrect labeling. In order to further improve the final quality, 8 senior inspectors are divided into two groups to conduct second-step verification. Each image was revisited by four assessors, and it will be put back into the pool to relabel if two or more experts are skeptical of the annotation. This process will not end until there is no images left in the pool. After careful verification, the overall annotation quality can be ensured. This assurance phase spends about 880 hours.
	
	\setcounter{table}{0}
	\setcounter{figure}{0}
	\section{Benchmarking of SODA-D}
	In this section, we elucidate the training details about SODA-D, also, the effects of label assignment and loss deigns to small object detection were also discussed.
	
	\subsection{Training Details}\label{app:sodad details}
	\textbf{Training hyperparameters}. Here we first illustrate some common settings in the benchmarking experiments of SODA-D. The default optimizer is Stochastic Gradient Descent (SGD) while except for Sparse RCNN \cite{154} and Deformable DETR \cite{52} that are optimized with AdamW. Most of model is trained for $1\times$ schedule that comprises a budget of 12 epochs, and the learning rate is decayed by 10 at 9 and 12 epoch, respectively. The weight decay is set to $0.0001$ for all baseline detectors, and we use Warmup technique to stabilize the initial training process, specifically, the learning rate will increase linearly to reach the predefined initial learning rate at first several iterations. Moreover, the images in SODA-D enjoy very high resolutions thereby directly feeding them into deep model is infeasible due to the GPU memory limitation. To overcome this issue we first crop the high-resolution images into a series of patches to perform detection, then the patch-wise results will be mapped to original images and  on which the (image-wise) NMS operation was conducted. In the process, the IoU thresholds of  patch-wise NMS and image-wise NMS are both set to $0.5$.
	
	\textbf{Model-wise settings}. To illustrate the detailed settings of the baseline detectors, we exhibit the learning rate and the pyramidal features of training each model in Tab. \ref{tab:TabB1}. Noting that CornerNet \cite{51}, CenterNet \cite{47}, Deformable-DETR \cite{52} and YOLOX \cite{YOLOX} design tailored neck to obtain high-quality multi-level representations, hence we do not include them. Specific architecture designs of each model please refer to the corresponding configurations in our codes which are available at \url{https://shaunyuan22.github.io/SODA}.

	\begin{table}[h]
		\centering
		\vspace{-1em}
		\caption{Detailed settings of each baseline detector for SODA-D. Note that the learning rate is set when the batch size is 8, and $P_2,P_3,P_4,P_5, P_6$ in pyramidal levels correspond to the feature strides of $4,8,16,32,64$. }
		\vspace{-1em}
		\resizebox{\columnwidth}{24mm}{
			\begin{tabular}{|c|c|c|}
				\hline
				Method & Learning Rate & Pyramidal Features \\
				\hline
				\hline
				Faster RCNN \cite{1} & 0.02 & $ P_2,P_3,P_4,P_5,P_6 $ \\
				Cascade RCNN \cite{151} & 0.02 & $ P_2,P_3,P_4,P_5,P_6 $ \\
				RetianNet \cite{3} & 0.01 & $ P_3,P_4,P_5,P_6 $ \\
				CornerNet \cite{51}  & 0.0000833 & / \\
				CenterNet \cite{47}  & 0.025 & / \\
				FCOS \cite{4}  & 0.005 & $ P_3,P_4,P_5,P_6 $ \\
				RepPoints \cite{152} & 0.01 & $ P_3,P_4,P_5,P_6 $ \\
				ATSS \cite{153}  & 0.01 & $ P_3,P_4,P_5,P_6 $ \\
				Deformable-DETR \cite{52} & 0.000025 & / \\
				Sparse RCNN \cite{154} & 0.0000125 & $ P_2,P_3,P_4,P_5,P_6 $ \\
				YOLOX \cite{YOLOX}  & 0.0001563 & / \\
				RFLA \cite{137}  & 0.02 & $ P_2,P_3,P_4,P_5,P_6 $ \\
				\hline
			\end{tabular}%
			\label{tab:TabB1}%
		}
	\end{table}%
	
	\subsection{The Effect of Label Assignment}\label{app:label assignmen}
	As we alluded to before, the label assignment strategy plays a significant role in training a deep detector, hence we will discuss the effect of label assignment to small object detection in this section. Concretely, we take Faster RCNN \cite{1}, RetinaNet \cite{3} and FCOS \cite{4} as baselines or reference methods to investigate the performances of different strategies.
	
	\begin{table*}[t]
		\scriptsize
		\centering
		\caption{The effect of label assignment strategies to SOD. All the approaches take ResNet-50 as the backbone and trained for '1$\times$' schedule.}
		\vspace{-1em}
		\resizebox{0.75\textwidth}{16.65mm}{
			\begin{tabular}{|c|ccc|cccc|c|c|}
				\hline
				Method & $AP$  & $AP_{50}$  & $AP_{75}$  & $AP_{eS}$  & $AP_{rS}$  & $AP_{gS}$  & $AP_{N}$  & $\#$Param. & FLOPs \\
				\hline
				\hline
				Faster RCNN \cite{1} & 28.9 & 59.7 & 24.2 & 13.9 & 25.6 & 34.3  & 43.2  & 41.16M & 292.28G  \\
				RFLA \cite{137} &  29.7  & 60.2  & 25.2  & 13.2  & 26.9  & 35.4  & 44.6  & 41.16M & 292.06G  \\
				\hline
				\hline
				RetianNet \cite{3} & 28.2  & 57.6 & 23.7 & 11.9    & 25.2  & 34.1  & 44.2  & 35.68M & 299.5G  \\
				FreeAnchor \cite{freeanchor}  & 29.6 & 58.4 & 25.6 & 13.3 & 26.7 & 35.5 & 45.5 & 35.68M & 295.5G \\
				PAA \cite{PAA}  & 29.7 & 56.9 & 26.5 & 12.0 & 26.3 & 36.3 & 46.3 & 31.32M & 290.79G \\
				\hline
				\hline
				FCOS \cite{4}  & 23.9  & 49.5  & 19.9    & 6.9  & 19.4  & 30.9  & 40.6  & 31.86M & 284.53G  \\
				ATSS \cite{153} & 26.8 & 55.6 & 22.1 & 11.7 & 23.9 & 32.2 & 41.3  & 31.32M  & 290.79G \\
				AutoAssign \cite{autoassign}  & 25.5 & 52.4 & 21.6 & 9.6 & 21.9 & 31.7 & 41.0 & 35.4M & 285.48G \\
				\hline
			\end{tabular}%
			\label{tab:TabB2}%
		}
		\vspace{-0.5em}
	\end{table*}%

	From Tab. \ref{tab:TabB2}, RFLA \cite{137} can boost the overall $AP$ of Faster RCNN \cite{1} by $0.8\%$ points but actually deteriorates $AP_{eS}$. We conjecture that the Gaussian Receptive Field-based scheme cannot assign adequate samples for \textit{extremely Small} instances, because there is only one prior when calculating the distances between the gaussian prior and ground-truth objects. 
	By modeling the training procedure as a maximum likelihood estimation (MLE) problem, FreeAnchor \cite{freeanchor} frees the hand-crafted anchor matching strategy and achieves comprehensive improvements compared to RetinaNet. Moreover, FreeAnchor integrates the recall rate into the optimization process which guarantees those size-limited objects could obtain at least one prediction.
	PAA \cite{PAA} considers classification and localization both in label assignment, optimization and post-processing, which is different from the previous works. Specifically, they formulate the anchor assignment as a probabilistic procedure by calculating anchor scores from a detector model and maximizing the likelihood of these scores for a probability distribution. If we delve into the specific metrics, though the $AP$ of PAA is higher than that of RetinaNet, the $AP_{eS}$ actually does not grows and the $AP$ of larger instances was improved substantially. We deem that PAA struggles to capture the distribution of small objects from the scores of limited priors (single anchor box per location) and the intrinsic difficulty of small objects. Moreover, thanks to the consideration of location during the whole training process, the $AP_{75}$ of PAA is much higher than its competitors, which, in other words, reveals that the primary problem of small objects lies in the missing detection. 
	For anchor-free method FCOS, ATSS \cite{153} significantly improves the baseline performance especially for $AP_{eS}$, showing that the dynamic assignment scheme is robust and conducive for objects of all scales in SODA-D. AutoAssign \cite{autoassign} deems that the points inside the object regions should not be treated as positive because only a part of pixels in ground-truth box belong to foreground, in view of this, they design a re-weighting strategy to adjust the \textit{pos/neg} assignment of each instance. AutoAssign can achieve $26.8 \% AP$, but when compared to ATSS, the performance gain is limited ($+1.6\%$ \textit{v.s.} $+2.9\%$) and this may caused by the blurred appearance of small objects. 
	
	In summary, it can be noticed from the aforementioned results that the prevailing label assignment strategies seem cannot handle the instances who have extremely limited sizes well, but for large objects, these schemes can boost the performance substantially. Moreover, the paradigms based on densely arranged priors still have predominance in comparison to their competitors.
	
	\subsection{The Effect of Loss Function}\label{app:loss function}
	In this section, we take RetinaNet and ATSS as the baseline and reference model to investigate how the loss designs can affect the performance of detectors on small objects. And considering there have no tailored loss functions for SOD task, we take GHM \cite{GHM} and GFL \cite{GFL}, which have been proven effective on generic object detection, to conduct the experiments and the results are shown in Tab. \ref{tab:TabB3}.
	
	\begin{table}[h]
		\scriptsize
		\centering
		\caption{The effect of loss designs to SOD. All the approaches take ResNet-50 as the backbone and trained for '1$\times$' schedule.}
		\vspace{-1em}
		\resizebox{\columnwidth}{9.25mm}{
			\begin{tabular}{|c|ccc|cccc|}
				\hline
				Method & $AP$  & $AP_{50}$  & $AP_{75}$  & $AP_{eS}$  & $AP_{rS}$  & $AP_{gS}$  & $AP_{N}$ \\
				\hline
				\hline
				RetianNet \cite{3} & 28.2  & 57.6 & 23.7 & 11.9 & 25.2  & 34.1  & 44.2 \\
				GHM \cite{GHM}  & 28.4 & 57.7 & 23.9 & 12.5 & 25.6 & 34.0 & 43.8 \\
				\hline
				\hline
				ATSS \cite{153} & 26.8 & 55.6 & 22.1 & 11.7 & 23.9 & 32.2 & 41.3  \\ 
				GFL \cite{GFL}  & 29.0 & 57.3 & 25.2 & 12.8 & 25.4 & 35.1 & 44.2 \\
				\hline
			\end{tabular}%
			\label{tab:TabB3}%
		}
		\vspace{-0.5em}
	\end{table}%
	
	GHM assumes that the vanilla Focal Loss can alleviate the imbalance issue but meanwhile may pay much attention to fit the outliers, which are detrimental to the overall training procedure. Hence they propose to decay the weight of those samples that the model cannot deal with. GHM obtains an overall $AP$ with $28.4 \% $ points which is slightly ahead of RetinaNet and surprisingly, it can outperform RetinaNet with $0.6 \% $ points on $AP_{eS}$. We speculate that this is because the objects of \textit{extremely Small} category are usually with limited information and distorted structures even cannot be recognized, as discussed in AdaFace \cite{AdaFace}. 
	GFL reconciles the optimization between classification and centerness score during training, and furthermore, they model the representation of a bounding box as General distribution instead of common Dirac delta distribution to dispose the detection under complex scenes. From Tab. \ref{tab:TabB3}, GFL surpasses ATSS by a substantial margin at all metrics, this can be largely attributed to the remarkable capability about capturing the uncertain boundaries of small instances, the remarkable  $AP_{75}$ offers further grounds.

	\setcounter{table}{0}
	\setcounter{figure}{0}
	\section{Benchmarking of SODA-A}
	In this section, we elucidate the training details about SODA-A, and the settings about the proposal parameters were also discussed.
	
	\subsection{Training Details}\label{app:sodaa details}
	\textbf{Training hyperparameters}. The commonly used hyperparamteters when training the baseline approaches of SODA-A are similar to that of SODA-D, the only difference lies in that the IoU threshold of patch-wise NMS operation is set to 0.1 which is in accordance with the default setting of mmrotate \cite{157}.
	
	\textbf{Model-wise settings}. To illustrate the detailed settings of these baseline detectors of SODA-A, we exhibit the learning rate as well as the pyramidal features of training each model in Tab. \ref{tab:TabC1}. Specific architecture designs of each model please refer to the corresponding configurations in our codes which are available at \url{https://shaunyuan22.github.io/SODA}.
	
	\begin{table}[h]
		\centering
		\vspace{-1em}
		\caption{Detailed settings of each baseline detector for SODA-A. Note that the learning rate is set when the batch size is 4, and $P_2,P_3,P_4,P_5, P_6$ in pyramidal levels correspond to the feature strides of $4,8,16,32,64$. }
		\vspace{-1em}
		\resizebox{\columnwidth}{18.5mm}{
			\begin{tabular}{|c|c|c|}
				\hline
				Method & Learning Rate & Pyramidal Features \\
				\hline
				\hline
				Rotated Faster RCNN \cite{1} & 0.01 & \multirow{9}[0]{*}{$ P_2,P_3,P_4,P_5,P_6 $} \\
				Rotated RetinaNet \cite{3} & 0.005 & \\
				RoI Transformer \cite{159} & 0.01 & \\
				Gliding Vertex \cite{160}  & 0.01 & \\
				Oriented RCNN \cite{161}  & 0.01 & \\
				S$^2$A-Net \cite{162}  & 0.005 & \\
				DODet \cite{163} & 0.01 & \\
				Oriented RepPoints \cite{ORep}  & 0.016 & \\
				DHRec \cite{DHRec}  & 0.005 & \\
				\hline
			\end{tabular}%
			\label{tab:TabC1}%
		}
	\end{table}%

	\subsection{Number of Proposals}\label{app:proposals}
	As we alluded to before, the objects in our SODA-A may distribute in a very dense fashion, which actually requires deliberate settings about proposal parameters. Intuitively, we have to strike a balance between proposal numbers and detection accuracy, in other words, computational consumption and accuracy. Excessive proposals could ensure the recall rate, though, it involves massive computation concurrently. While inadequate proposals hinder the overall performance. To determine the optimal choice about patch-level proposal numbers for best performance on SODA-A, we train Oriented RCNN (with a ResNet-50 \cite{10} as backbone network) with train-set and test on the test-set. Tab. \ref{tab:TabC2} reports the results with different proposal number settings. We can see that the $AP$ and $AP_{T}$ performance vary slightly when the proposal numbers change from 2000 to 8000, whereas the detection speed decreases dramatically. Hence we set the proposal number to 2000 for optimal performance, both accuracy and speed, in our experiments.
	
	\begin{table}[h]
		\centering
		\vspace{-1em}
		\caption{$AP$ \textit{vs.} \textit{Speed} of Oriented RCNN \cite{161} with different number of proposals per patch on SODA-A test-set. FPS is tested on a single RTX 2080Ti GPU.}
		% \caption{$AP$ of Oriented RCNN \cite{161} with different number of proposals per patch on SODA-A test-set.}
		\vspace{-1em}
		\resizebox{\columnwidth}{6mm}{
			\begin{tabular}{|c|cccccccc|}
				\hline
				Proposal Num. & 1000  & 2000  & 3000  & 4000  & 5000  & 6000  & 7000  & 8000 \\
				\hline
				\hline
				$AP$   & 33.9  & 34.4  & 34.5  & 34.4  & 34.4  & 34.6  & 34.2  & 34.1 \\
				\textit{Speed} (FPS) & 13.3 & 12.3 & 11.0 & 10.5 & 9.9 & 9.4 & 9.1 & 8.7 \\
				\hline
			\end{tabular}%
			\label{tab:TabC2}%
		}
		\vspace{-1em}
	\end{table}%

\end{appendices}

%\bibliographystyle{IEEEtran}
%\bibliography{SODA_app}

\end{document}